\documentclass{article}
\PassOptionsToPackage{numbers,sort&compress}{natbib}
\usepackage{decompgym}
\usepackage[utf8]{inputenc}
\usepackage{amsmath,amssymb,amsfonts}
\usepackage{graphicx,booktabs,tabularx,array}
\usepackage{multirow}
\usepackage{xcolor,microtype,url}
\usepackage{enumitem}
\usepackage{placeins,needspace}
\usepackage{etoolbox}
\usepackage{hyperref}
\definecolor{paperlinkblue}{HTML}{000080}
\hypersetup{colorlinks=true,allcolors=paperlinkblue,allbordercolors=paperlinkblue,pdflinkmargin=0pt}
\AddToHook{begindocument/end}{\hypersetup{pdfborder={0 0 1},pdfborderstyle={/S/U/W 0.3}}}
\makeatletter
\patchcmd{\Hy@StartlinkName}{\pdfstartlink}{\pdfstartlink depth 1.5pt}{}{\PackageError{paper_setup}{Cannot set internal-link underline depth}{Compile with pdfLaTeX and check the hyperref driver.}}
\patchcmd{\hyper@linkurl}{\pdfstartlink}{\pdfstartlink depth 1.5pt}{}{\PackageError{paper_setup}{Cannot set URL underline depth}{Compile with pdfLaTeX and check the hyperref driver.}}
\makeatother
\usepackage[capitalise,nameinlink,noabbrev]{cleveref}
\crefname{figure}{Figure}{Figures}
\crefname{table}{Table}{Tables}
\crefname{section}{Section}{Sections}
\crefname{subsection}{Section}{Sections}
\crefname{subsubsection}{Section}{Sections}
\crefname{appendix}{Appendix}{Appendices}
\crefname{equation}{Equation}{Equations}
\AddToHook{cmd/appendix/after}{\crefalias{section}{appendix}\crefalias{subsection}{appendix}\crefalias{subsubsection}{appendix}}
\usepackage{listings}
\usepackage[varqu,scaled=0.95]{zi4}
\usepackage[most]{tcolorbox}
\definecolor{promptframe}{HTML}{4A5A78}
\definecolor{promptback}{HTML}{F7F8FB}
\definecolor{promptmark}{HTML}{B0251A}
\lstdefinestyle{prompt}{basicstyle=\ttfamily\scriptsize,commentstyle=\color{black!55}\upshape,keywordstyle=,stringstyle=,lineskip=-0.5pt,aboveskip=0pt,belowskip=0pt,breaklines=true,breakatwhitespace=true,breakindent=0pt,columns=fullflexible,keepspaces=true,showstringspaces=false,upquote=true,moredelim=[is][\color{promptmark}\bfseries]{@@}{@@},moredelim=[is][\color{black!50}]{<<}{>>}}
\newtcbinputlisting[auto counter,crefname={Prompt}{Prompts}]{\promptfilebox}[3]{listing file={prompts/#2},listing only,listing options={style=prompt},enhanced,breakable,colback=promptback,colframe=promptframe,boxrule=0.6pt,arc=2pt,left=5pt,right=5pt,top=4pt,bottom=4pt,fonttitle=\small\bfseries,coltitle=white,title={Prompt~\thetcbcounter: #1},label={#3}}
\newcommand{\promptfile}[3]{\Needspace{12\baselineskip}\promptfilebox{#1}{#2}{#3}}
\newcommand{\skeleton}[2]{}
\newcommand{\code}[1]{\texttt{#1}}
\title{DecompGym: Benchmarking LLM Agents on Designing Decomposition Algorithms}
\authors{Sikai Cheng$^{*\dagger}$, Tinghan Ye$^{*}$, Arnaud Deza, Xiaonan Qi, Miao Li, Litong Liu, Changkun Guan}
\affiliations{Georgia Institute of Technology\hspace{1.5em}$^{*}$Equal contribution\hspace{1.5em}$^{\dagger}$Corresponding author}
\contactemail{sikaicheng@gatech.edu}
\date{\today}
\begin{document}
\maketitle
\begin{abstract}
Benders decomposition and column generation solve large optimization problems by passing cuts, columns, and dual values between a master problem and subproblems, and are hard to implement correctly. Large language models now automate parts of algorithm design, which points toward agents that build decomposition solvers for new problems. As a first step toward such agents, we study how well current coding agents design and implement these two algorithms, and how they fail. We introduce DecompGym, 81 instances from nine problem families (five Benders, four column generation), in which an agent writes a solver for a prescribed method. Each solver must meet three criteria: execution within a 60-second time limit; numerical feasibility together with a method audit, in which an LLM judge reads the solver's code; and an objective within 1\% of the stored optimal objective. We evaluate 12 GPT, Claude, and Gemini models running in coding-agent tools. Over up to ten attempts, we compare independent regeneration with reflection, which first diagnoses the failure. On the first attempt, two models solve every instance and seven solve more than 90\%. Most Benders failures (48 of 62) are missing or suboptimal solutions. The most common column-generation failure is a method-audit rejection. Output checks alone would accept 93 of the 108 attempts that fail the method audit. The most common rejection, an optimality claim made without proof, occurs 96 times, 92 of them in one model. Reflection turns 66\% of retries into successes, compared with 39\% for regeneration, and it repairs method-audit failures, for which the judge names the defect, far more often than objective gaps. Agents for decomposition therefore need checks on how a solver reaches its answer and feedback that locates the error. The benchmark is available at \url{https://github.com/SIKAI-C/DecompGym}.
\end{abstract}
\begin{figure}[H]
\centering
\includegraphics[width=\linewidth]{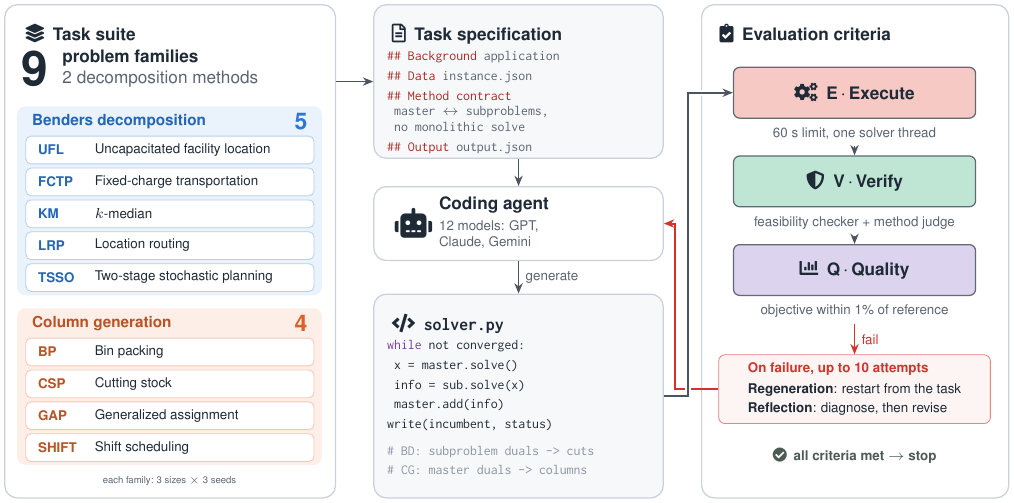}
\caption{DecompGym overview. A coding agent writes a solver from the task inputs; the solver is checked against three criteria in order (execution E, verification V, quality Q); failed attempts are retried by regeneration or by reflection.}
\label{fig:overview}
\end{figure}
\section{Introduction}
\label{sec:intro}

Many large optimization problems in operations research are solved by decomposition. Benders decomposition fixes complicating decisions in a master problem and adds cuts generated by subproblems~\citep{benders1962partitioning,rahmaniani2017benders}. Column generation solves a restricted master problem over a subset of columns and adds columns found by a pricing subproblem that uses the master's dual values~\citep{dantzig1960decomposition,lubbecke2005selected}. Both algorithms are difficult to implement. The master and subproblems must be formulated consistently; duals, cuts, or columns must pass correctly between them; and the stopping rule must match what the final solution certifies. An error in any of these steps can leave a program that runs and returns a plausible answer.

Large language models are now used to automate parts of algorithm design. Program-search methods evolve heuristics and algorithms with an LLM in the loop~\citep{romeraparedes2023mathematical,liu2024evolution,ye2024reevo,novikov2025alphaevolve}. In operations research, LLMs formulate optimization models~\citep{yang2025optibench,chen2026optengine}, design algorithms that match or beat published methods on most instances of inventory, queueing, and assortment problems~\citep{baek2026algorithms}, search over inventory policy classes~\citep{yang2026inventory}, and act as preference oracles inside multi-objective selection algorithms~\citep{jovine2026listen}. Recent benchmarks evaluate coding agents on writing complete solvers for operations research tasks~\citep{kong2026frontieror,li2026oragentbench}. These advances point toward agents that design and implement decomposition algorithms for new problems.

Building such agents first requires knowing how well current coding agents design and implement these algorithms and how they fail. This paper is an exploratory study of both. Existing evaluations let the agent choose the method and score the solution or policy it returns. Studies of coding agents show that passing tests does not imply meeting the specification when shortcuts are available~\citep{zhong2025impossiblebench,zhao2026specbench,gabor2025evilgenie}. For decomposition, output checks credit two kinds of solver that violate the method contract: one that reaches the optimal objective through a heuristic that bypasses the decomposition, and one that labels its solution optimal although its algorithm has not proved it. Measuring method use therefore requires an audit of the solver's code. We define the \emph{method-conformance gap} as the attempts that pass every output check (execution, feasibility, and objective within 1\% of the optimal objective) and fail the method audit. We measure it on ordinary tasks that plant no shortcut.

We conduct the study on DecompGym, a set of 81 instances from nine problem families (\cref{fig:overview}) drawn from facility location, transportation, routing, stochastic planning, packing, assignment, and scheduling. Five families are posed for Benders decomposition and four for column generation, each at three sizes with three seeds. An agent receives an application description, instance data, an output format, and a method contract that specifies the required master--subproblem information flow, then writes a standalone Python solver. Each attempt is checked against three criteria in sequence: execution within a common 60-second time limit; numerical feasibility checks together with a method audit, in which an LLM judge reads the solver's code; and an objective within 1\% of the stored optimal objective. A paired protocol shares the first attempt between two retry policies, independent regeneration and reflection (a separate model query that diagnoses the failed attempt and writes advice for the next one), so both start from the same outcome. The protocol records the attempts, time, and tokens each policy uses. DecompGym, including its instance generators, formulations, reference implementations, reference solutions, and the evaluation code, is available at \url{https://github.com/SIKAI-C/DecompGym}.

\subsection{Main findings}
\label{sec:findings}

We evaluate 12 models: five GPT, six Claude, and one Gemini. Each runs inside its provider's coding-agent tool, so its results reflect both the model and the tool. The main findings are as follows.
\begin{itemize}[leftmargin=1.4em,itemsep=1pt,topsep=2pt]
\item \textbf{Two models implement verified decompositions on every first attempt.} GPT-6 Astra and GPT-6.1 Sol pass every instance on their first attempt, and seven models exceed 90\%. First-attempt success ranges from 56.8\% to 100\%. With ten attempts, nine of the twelve models reach full coverage without reflection and ten with it.
\item \textbf{Execution overstates capability, and the failure modes depend on the method.} Of the first attempts, 99.0\% run and 87.8\% meet all three criteria. Most Benders failures (48 of 62) are missing or suboptimal solutions, concentrated in fixed-charge transportation. The most common column-generation failure is a method-audit rejection, in 26 of the 57 failed column-generation first attempts.
\item \textbf{The method-conformance gap is 93 of the 108 attempts rejected by the method audit.} These attempts returned feasible solutions within 1\% of the optimal objective, so an evaluation that checks only outputs would credit them. Of the 108, 96 reported an optimal solution without proof, and Claude Sonnet 5 produced 92 of them. In 95 of the 96, a column-generation solver had optimized only over the columns it generated and then claimed that no better solution exists. Counting these rejections as passes would raise Claude Sonnet 5's first-attempt successes from 46 to 66 of 81, and no other model's by more than one instance.
\item \textbf{Reflection succeeds most on failures that its feedback names.} Reflection turns 66\% of retries into successes, compared with 39\% for regeneration, and the method audit rejects 5 of its retries, compared with 64 under regeneration. Its remaining failures are mostly objective gaps, for which the feedback only reports that the target is unmet.
\item \textbf{Time and tokens measure different costs.} Among models that solve every instance without reflection, mean agent time to success ranges from 21.5 to 228.2 seconds. Token use ranks models differently. GPT-6 Luna and GPT-5.6 Luna finish faster than GPT-6 Astra and GPT-6.1 Sol but consume 2.2 to 3.2 times as many tokens.
\end{itemize}

\subsection{Implications for decomposition agents}
\label{sec:implications}

An agent built for decomposition needs a check on how its solver reaches an answer, because an agent that tests only its output would accept every attempt in the method-conformance gap. It also needs feedback that locates the error. The judge's rationale locates method-audit failures, and after such a rejection the next reflection attempt met all three criteria in 41 of 44 cases. After an objective gap, the feedback gives no such location. Information from the optimization itself, such as the solver's bounds or its gap at termination, could tell the agent where the error lies. \cref{sec:conclusion} states the limits of the study.

\section{Related Work}
\label{sec:related}

\paragraph{Optimization modeling benchmarks.}
Most benchmarks for LLMs in optimization test modeling. The model turns a natural-language problem into a formulation, a general-purpose solver solves it, and the benchmark compares the objective or status with a reference. The datasets range from linear-programming word problems to industrial, nonlinear, and MIPLIB-derived instances~\citep{ramamonjison2023nl4opt,huang2025mamo,huang2025orlm,lu2025optmath,yang2025optibench,chen2026optengine,li2026constructing}, and a survey reports high error rates in their reference answers~\citep{xiao2025survey}. Systems built for these benchmarks produce the model and its code with agent teams, fine-tuning, or solver-verified rewards~\citep{ahmaditeshnizi2024optimus,xiao2024chain,jiang2025llmopt,chen2025solver}. Beyond building new models, LLMs also turn natural-language change requests into updates of a deployed model and choose how to re-optimize it~\citep{ye2026reopt}. Because a general-purpose solver supplies the algorithm, these tasks never ask the model to implement one, and the output alone decides success.

\paragraph{Agents that write optimization algorithms and solver code.}
When agents write the solution method itself, benchmarks score complete programs by feasibility, objective, and running time on large operations research tasks~\citep{kong2026frontieror,li2026oragentbench} or by agreement between the stated formulation and the code~\citep{fu2026optiagentbench}. Others have agents write solver components such as constraint handlers, cut selectors, and branching rules~\citep{xu2026agentic,nie2026milpevo}, or heuristics scored by solution quality~\citep{sun2025cobench,chen2026heurigym,imajuku2025alebench}, and algorithm-design studies search over programs or policy classes with an evaluator~\citep{romeraparedes2023mathematical,liu2024evolution,novikov2025alphaevolve,ye2024reevo,baek2026algorithms,yang2026inventory}. In all of this work the agent chooses the method and the output or its performance decides success. FrontierOR observes that stronger models more often choose decomposition or hybrid methods, and ORAgentBench notes that it lacks a process-level audit~\citep{kong2026frontieror,li2026oragentbench}. Benchmarks that do prescribe a method come from machine-learning research code, where PaperBench and ResearchCodeBench grade implementations of a described method with rubrics and unit tests~\citep{starace2025paperbench,hua2025researchcodebench}. DecompGym brings this kind of evaluation to optimization algorithms by prescribing the decomposition and checking the solver's code against it.

\paragraph{Decomposition methods and their automation.}
Decades of work have refined both methods. Benders variants strengthen the cuts or extend the method to nonlinear, logic-based, and big-M subproblems~\citep{magnanti1981accelerating,geoffrion1972generalized,hooker2003logic,codato2006combinatorial}, as surveyed by \citet{rahmaniani2017benders}. Column-generation research covers bounds, stabilization, branching, and integer solutions~\citep{lubbecke2005selected}, and branch-and-price embeds column generation in branch-and-bound~\citep{barnhart1998branch}. Solvers and learned models automate parts of applying a decomposition. GCG, SCIP, and CPLEX run a Dantzig--Wolfe or Benders decomposition when the user supplies the structure or the solver detects it~\citep{gamrath2010experiments,bergner2015automatic,maher2021implementing,bonami2020implementing}, and learned models decide whether to decompose or which columns to add~\citep{kruber2017learning,morabit2021machine,bengio2021machine}. These tools need a model whose block structure is given or detectable, and they run a fixed implementation. An agent in DecompGym receives a problem description and must write the master, the subproblems, and their coordination.

\paragraph{Specification conformance, false success, and judge validity.}
Coding-agent benchmarks score software changes with tests~\citep{jimenez2024swebench}, and several studies show that agents can pass tests without meeting the specification. The evidence comes from tasks in which the tests conflict with the specification or held-out tests differ from visible ones~\citep{zhong2025impossiblebench,zhao2026specbench}, from measured rates of hardcoding, test editing, and reward hacking~\citep{gabor2025evilgenie,vonarx2025recent,thaman2026reward,huang2026reward}, and from agents' status reports, which are often false~\citep{advani2026confident,smyth2026quantifying}. Closest to our setting, LLM-written combinatorial solvers replace complete search with approximations or unverified bounds~\citep{wang2026formalize}. Most of these studies plant a shortcut or make one available, and they do not measure how often output checks miss method failures on ordinary optimization tasks. Measuring this rate requires a judge. LLM judges agree with humans on most comparisons but show position, verbosity, and self-preference biases~\citep{zheng2023judging,panickssery2024llm}; they detect false success poorly from agent trajectories and reliably from clear code-level cheating~\citep{advani2026confident,gabor2025evilgenie}. The DecompGym judge reads solver source, which is closer to the code-level setting.

\paragraph{Reflection and repair.}
Reflection methods give an agent verbal feedback on its earlier attempts before it tries again~\citep{shinn2023reflexion,madaan2023self,chen2023teaching}. Their benefit depends on where the feedback comes from and how cost is counted. Models do not improve their reasoning through self-correction without external feedback~\citep{huang2023large}, self-repair gains in code generation are modest once the repair step is charged~\citep{olausson2023self}, and repeated sampling with a verifier raises coverage log-linearly in the number of samples~\citep{brown2024large}. In DecompGym, regeneration is repeated sampling with the benchmark's evaluation as verifier, and reflection receives external feedback from execution and the method audit. Both policies start from the same first attempt, and the benchmark records the attempts, time, and tokens each uses.

\section{The DecompGym Benchmark}
\label{sec:problem}

Each DecompGym task gives a coding agent one instance of an optimization problem and a prescribed decomposition method, and asks for a solver that applies that method.

A task consists of an application description, the instance data in a structured file, a prescribed method family, and an output contract. The agent writes a Python solver that reads the instance and outputs an objective value and the decisions needed for verification. The permitted libraries are Gurobi's Python interface, NumPy, and the Python standard library. The agent designs the decomposition from this specification and writes all of the solver's code.

\begin{table}[!b]
\centering
\caption{Nine task families, with nine instances per family. BD denotes Benders decomposition; CG denotes column generation. The decomposition-structure column summarizes what the master decides and what the subproblems return to it; \cref{app:formulations} states the formulations.}
\label{tab:families}
\small
\begin{tabularx}{\linewidth}{llX}
\toprule
Method & Family & Decomposition structure \\
\midrule
\multirow{5}{*}{BD} & UFL & Facility-opening decisions fix assignment subproblems, which return optimality cuts. \\
 & FCTP & Arc-opening decisions fix a transportation subproblem, which returns optimality or feasibility cuts. \\
 & KM & Location-selection decisions fix an assignment subproblem, which returns optimality cuts. \\
 & LRP & Depot-opening decisions fix route selection, whose LP relaxation returns optimality or feasibility cuts. \\
 & TSSO & Capacity-activation decisions fix the scenario planning subproblems, which return optimality cuts. \\
\midrule
\multirow{4}{*}{CG} & BP & The master weights packing patterns to cover items; a knapsack over item sizes returns new patterns. \\
 & CSP & The master weights cutting patterns to meet demands; a knapsack over piece widths returns new patterns. \\
 & GAP & The master assigns each job once via per-machine job subsets; per-machine knapsacks return new subsets. \\
 & SHIFT & The master weights shifts to cover time-slot demands; pricing over legal intervals returns new shifts. \\
\bottomrule
\end{tabularx}
\end{table}

\cref{tab:families} lists the nine families. Five are posed for Benders decomposition and four for column generation, so each method is exercised on several distinct master--subproblem structures. Among the Benders families, uncapacitated facility location (UFL) has a well-studied Benders decomposition~\citep{fischetti2017redesigning}, fixed-charge transportation (FCTP) is a fixed-charge network design problem, a class to which Benders decomposition is widely applied~\citep{costa2005survey}, $k$-median (KM) admits an efficient Benders decomposition~\citep{duranmateluna2023efficient}, and stochastic planning (TSSO) follows the two-stage L-shaped structure~\citep{vanslyke1969lshaped,laporte1993integer}. Location routing (LRP) combines depot location with route selection~\citep{prodhon2014survey}; its route-selection subproblem is integer, so linear-programming cuts give only a bound, and closing the gap calls for logic-based Benders cuts~\citep{hooker2003logic,fazelzarandi2012using}. Among the column-generation families, cutting stock (CSP) is the original column-generation problem~\citep{gilmore1961linear}, generalized assignment (GAP) has a branch-and-price literature~\citep{savelsbergh1997branch}, shift scheduling (SHIFT) is solved by branch-and-price over a set-covering formulation~\citep{mehrotra2000optimal}, and bin packing (BP) and cutting stock are studied with branch-and-price and other exact methods~\citep{vance1998branch,delorme2016bin}. \cref{app:formulations} gives each family's model and the formulation of its reference decomposition. Each family is instantiated at three size levels with three seeds, for 45 Benders and 36 column-generation instances in total. Size levels are defined by the generator settings within each family. The instances are of moderate dimension, and every solver an agent writes must finish within 60 seconds; a hand-written reference implementation of each prescribed decomposition meets all three criteria within this limit (\cref{app:formulations}). The benchmark measures whether an agent implements the prescribed method correctly, and the time limit also requires that implementation to be reasonably efficient. Every model receives the same inputs, prompts, and output schemas and is scored against the same stored optimal objectives. \cref{app:inventory} lists the realized dimensions.

A Benders implementation must fix master decisions in the subproblems and feed subproblem information back to the master through feasibility or optimality cuts. Classical, integer, and logic-based variants are permitted~\citep{laporte1993integer,hooker2003logic}. A column-generation implementation must pass restricted-master dual information to pricing and use the resulting columns in subsequent master solves. Legitimate initialization, immediate convergence, and final integer restricted-master solves are allowed. A separate monolithic solve of the full original problem is prohibited anywhere in the solver, including the initial solution and any backup procedure. The contract also requires mutually consistent decisions and objective values, and a reported status that matches what the algorithm proves: the label \emph{optimal} requires a proof that no better solution exists, and any other solution must be labeled \emph{feasible}. The method contract states this rule in its closing sentence (\cref{prompt:generation}).

\section{Evaluation Protocol}
\label{sec:methodology}

Each of the twelve models attempts all 81 instances under two retry policies, and three criteria score every attempt in order.

The study evaluates five GPT models through OpenAI's Codex command-line tool, six Claude models through Claude Code, and Gemini 3.8 Flash through Google's Antigravity command-line tool, all at medium reasoning effort; \cref{tab:main-results} lists the twelve. Every agent call runs in a new Docker container with a fresh workspace, and no conversation history, memory, user configuration, or extensions carry over from earlier calls. Each solver the agent writes runs with one solver thread and a 60-second limit that covers setup, all master and subproblem solves, and writing the output. The method judge is a separate GPT-6 Astra query at high reasoning effort. \cref{app:configuration} lists software versions and limits.

The first attempt on each instance is shared by the two policies (\cref{fig:overview}). In the regeneration policy, each later attempt starts a new session from the original task and data, so each retry is independent of earlier attempts. In the reflection policy, a separate diagnosis call to the same model first reads the failed code, its execution output, and the judge's rationale and writes advice. The next attempt receives the previous code together with that advice. When the objective misses the target, the diagnosis prompt states only that the objective must improve. Each policy stops at the first success or after ten attempts.

Execution (E) requires the solver to run without error within the time limit and write its output file. Verification (V) adds two checks. The output must report a finite objective and pass problem-specific numerical checks of feasibility, integrality, consistency, and the recomputed objective, and the method judge must accept the code. The judge reads the solver's source and the method contract given to the agent, checks whether the code implements the prescribed decomposition and whether its iterative master--subproblem procedure actually produces the reported solution, and returns a verdict with a written rationale. Quality (Q) adds that the recomputed objective lies within 1\% of the stored optimal objective, which the agent never sees. An attempt succeeds when it meets all three criteria.

Results are reported by attempt budget. An instance counts as solved at budget $k$ (written Q@$k$ for the quality criterion) when one of its first $k$ attempts succeeds, and every rate keeps all 81 instances in the denominator at every budget. Pooled counts cover the 972 model--instance pairs. \cref{app:scoring} gives the formal definitions and the counting conventions.

Cost is measured by agent time and tokens. Agent time is the time the agent spends writing solvers and diagnosing failures, up to success or the last allowed attempt; judge time and solver run time are reported separately. Token counts follow each tool's own accounting. \cref{app:resources} breaks the recorded time down by stage and defines the token fields of each tool.

To locate failures, each attempt is assigned to the first check it fails: execution, a numerical check, the method audit, or the objective target. Each method rejection is classified by hand, from the judge's rationale, into one of three findings: an unproven optimality claim, an inactive decomposition, or a solution bypass (\cref{app:judge}).

\section{Results}
\label{sec:results}

\begin{table}[!htbp]
\centering
\caption{Quality success rates (\%). Q@$k$ counts an instance as solved when one of its first $k$ attempts meets all three criteria. P denotes independent regeneration and R reflection; Q@1 is shared because both policies use the same first attempt. Each model is evaluated on all 81 instances; the overall row pools 972 model--instance pairs. All models use medium reasoning effort.}
\label{tab:main-results}
\small
\begin{tabular}{lrrrrrrr}
\toprule
Model & Q@1 & Q@2 P & Q@2 R & Q@3 P & Q@3 R & Q@10 P & Q@10 R \\
\midrule
GPT-6 Astra & 100.0\% & 100.0\% & 100.0\% & 100.0\% & 100.0\% & 100.0\% & 100.0\% \\
GPT-6 Sol & 97.5\% & 98.8\% & 100.0\% & 100.0\% & 100.0\% & 100.0\% & 100.0\% \\
GPT-6.1 Sol & 100.0\% & 100.0\% & 100.0\% & 100.0\% & 100.0\% & 100.0\% & 100.0\% \\
GPT-6 Luna & 77.8\% & 88.9\% & 87.7\% & 93.8\% & 92.6\% & 95.1\% & 98.8\% \\
GPT-5.6 Luna & 67.9\% & 90.1\% & 93.8\% & 92.6\% & 100.0\% & 96.3\% & 100.0\% \\
Claude Opus 5.5 & 96.3\% & 100.0\% & 100.0\% & 100.0\% & 100.0\% & 100.0\% & 100.0\% \\
Claude Sonnet 5.5 & 95.1\% & 100.0\% & 100.0\% & 100.0\% & 100.0\% & 100.0\% & 100.0\% \\
Claude Haiku 5.5 & 87.7\% & 100.0\% & 98.8\% & 100.0\% & 100.0\% & 100.0\% & 100.0\% \\
Claude Opus 5 & 86.4\% & 95.1\% & 100.0\% & 96.3\% & 100.0\% & 100.0\% & 100.0\% \\
Claude Sonnet 5 & 56.8\% & 67.9\% & 88.9\% & 75.3\% & 96.3\% & 97.5\% & 98.8\% \\
Claude Fable 5 & 97.5\% & 100.0\% & 100.0\% & 100.0\% & 100.0\% & 100.0\% & 100.0\% \\
Gemini 3.8 Flash & 90.1\% & 95.1\% & 100.0\% & 97.5\% & 100.0\% & 100.0\% & 100.0\% \\
\textbf{Overall} & 87.8\% & 94.7\% & 97.4\% & 96.3\% & 99.1\% & 99.1\% & 99.8\% \\
\bottomrule
\end{tabular}

\end{table}

\subsection{Can agents design and implement decomposition algorithms?}
\label{sec:capability}

\begin{figure}[!htbp]
\centering
\includegraphics[width=.82\linewidth]{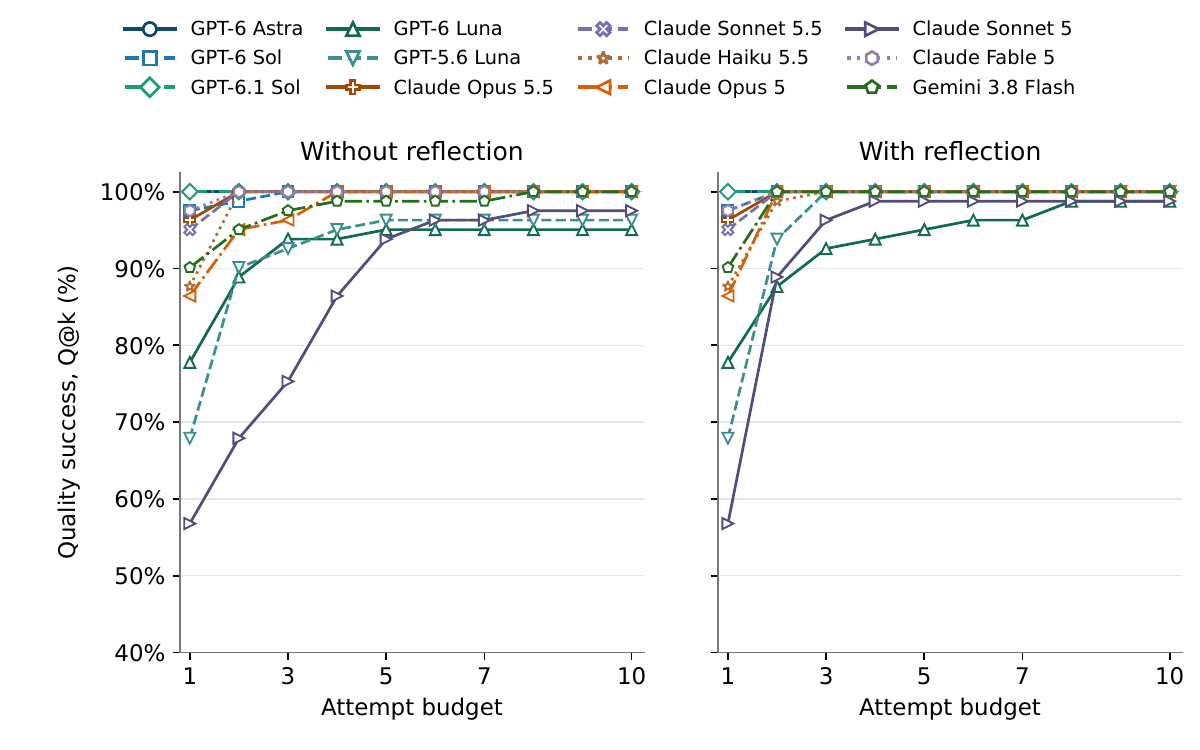}
\caption{Quality success by attempt budget, without (left) and with (right) reflection. Each curve keeps every instance in the denominator at every budget, and an instance solved early remains solved at later budgets. Both panels share the vertical scale, which starts at 40\%.}
\label{fig:attempt-quality}
\end{figure}

Seven of the twelve models exceed 90\% first-attempt quality success, and GPT-6 Astra and GPT-6.1 Sol meet all three criteria on every instance at their first attempt (\cref{tab:main-results}). Across all models, 853 of 972 first attempts (87.8\%) succeed. First-attempt success is lowest for Claude Sonnet 5 (56.8\%) and GPT-5.6 Luna (67.9\%). Retries recover 110 of the 119 failed pairs under regeneration and 117 under reflection. With up to ten attempts, nine models solve every instance without reflection and ten with reflection, and pooled success rises to 99.1\% and 99.8\%, respectively. \cref{fig:attempt-quality} shows that models with the same final coverage can need different numbers of attempts to reach it. Because final coverage is near saturation, first-attempt success and the cost of reaching a success separate the models.

\FloatBarrier
\subsection{Where do agents fail?}
\label{sec:failures}

\begin{figure}[!htbp]
\centering
\includegraphics[width=\linewidth]{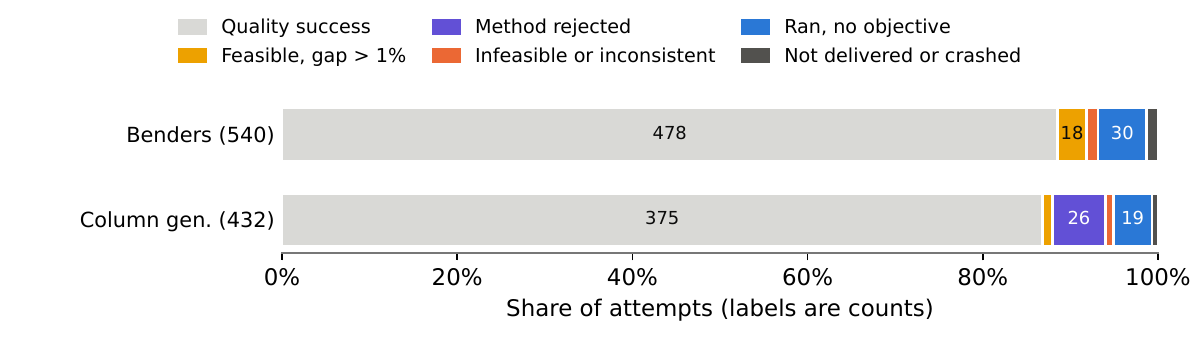}
\caption{First-attempt outcomes by the first check failed, split by decomposition method. Segment labels give counts; the smallest segments are unlabeled.}
\label{fig:failure-first}
\end{figure}

Of the 119 failed first attempts, 109 run to completion and fail a later criterion, so executable code is a weak signal of a correct decomposition. \cref{fig:failure-first} assigns each first attempt to the first check it fails and separates the two methods. Of the 62 Benders failures, 48 are missing or suboptimal solutions. Thirty attempts run without returning a finite objective, typically by reporting an error or declaring a feasible instance infeasible, and 18 return feasible solutions that miss the 1\% target. Fixed-charge transportation (FCTP) accounts for 19 of the 30 missing objectives and 9 of the 18 objective gaps. It is the hardest family, with 76 of its 108 first attempts succeeding, compared with 92 to 104 for every other family. GPT-6 Luna and GPT-5.6 Luna each solve only one of its nine instances on the first attempt. In column generation, the largest category, 26 of the 57 failed first attempts, has a feasible output but fails the method audit; seven further method rejections also fail a numerical check and are counted under that check. \cref{app:heterogeneity} breaks first-attempt success down by model and family, and \cref{app:stages} reports the cumulative execution, verification, and quality curves of every model.

The method audit catches failures that output checks cannot see. \cref{tab:judge-findings} groups all 108 rejections by finding. The most common finding, with 96 cases, is an \emph{unproven optimality claim}. In the typical case, a column-generation solver takes Gurobi's optimal status for the final integer restricted master as proof of global optimality. \cref{sec:case-study} works through an example and \cref{app:unproven} gives the argument. Eight rejections identify decomposition loops that never run or cannot influence the master. Four rejections identify a heuristic or full enumeration that produces the reported solution outside the decomposition; \cref{app:judge} gives examples of both. Of the 108 rejected attempts, 93 returned feasible solutions within 1\% of the optimal objective and would pass an output-only evaluation; these 93 attempts form the method-conformance gap.

One model accounts for nearly all unproven optimality claims. Claude Sonnet 5 makes 92 of the 96, and they account for 26 of its 30 first-attempt column-generation failures (\cref{tab:unproven-claims}). It passes 40 of 45 Benders instances on the first attempt but only 6 of 36 column-generation instances. No other model makes the claim more than twice across all of its attempts, so the requirement is attainable under the same prompt. Output checks alone would credit Claude Sonnet 5 with 20 more first-attempt successes (66 instead of 46 of 81; \cref{sec:robustness}), and only an audit of the code shows which model is affected. A benchmark that prescribes a method should verify it and report the method-conformance gap.

\begin{table}[!htbp]
\centering
\caption{Method-audit findings among the 1,429 physical attempts, from a manual reading of every judge rationale (\cref{app:judge}). Columns separate first attempts by method and retries by policy. Some rejected attempts also fail the numerical checks; \cref{fig:failure-first,fig:failure-retry} assign those to the numerical failure.}
\label{tab:judge-findings}
\small
\begin{tabular}{lrrrr}
\toprule
Judge finding & First BD & First CG & Retries P & Retries R \\
\midrule
Reports \code{optimal} without a proof of optimality & 0 & 29 & 63 & 4 \\
Decomposition loop never runs or cannot affect the master & 6 & 1 & 0 & 1 \\
Heuristic or enumeration fallback produces the solution & 0 & 3 & 1 & 0 \\
\midrule
All method rejections & 6 & 33 & 64 & 5 \\
\bottomrule
\end{tabular}

\end{table}

\FloatBarrier
\subsection{A case study: unproven optimality in column generation}
\label{sec:case-study}

On the medium cutting-stock instance with seed 0, Claude Sonnet 5's first attempt returns 87 stock pieces, which equals the optimal objective, and labels its solution optimal. It takes that label from Gurobi's status for the final integer restricted master, the integer program over the patterns generated so far. That status means only that no better integer solution exists using those patterns. Because the objective is integer, the solution is proved optimal only when $z_{\mathrm{IP}}(K)$, the value of the integer restricted master over the generated patterns $K$, equals $\lceil z_{\mathrm{LP}}\rceil$, where $z_{\mathrm{LP}}$ is the linear relaxation value after pricing converges (\cref{app:unproven} derives these bounds). The attempt passes the execution, feasibility, and objective checks. Its optimality claim may be true, but the code never compares the two bounds. The judge rejects it with the rationale reproduced in \cref{prompt:rationale}.

A conforming solver labels its solution feasible unless it can close this gap, and a feasible label is never penalized. After reading the rationale, the reflection step writes the advice in \cref{prompt:advice}. The reflection policy's second attempt returns the same 87 pieces and labels them optimal only if pricing has converged and $z_{\mathrm{IP}}(K)=\lceil z_{\mathrm{LP}}\rceil$, with $z_{\mathrm{LP}}$ taken from the last linear relaxation. The judge accepts it. The two attempts return identical outputs, and only the second attempt's code proves the optimality it reports. Claude Sonnet 5.5 and GPT-6 Astra perform this bound check in their first attempts on this instance. Without reflection, Claude Sonnet 5 repeats the first attempt's status rule in all ten attempts.

\FloatBarrier
\subsection{When does reflection help?}
\label{sec:reflection}

\begin{figure}[!htbp]
\centering
\includegraphics[width=\linewidth]{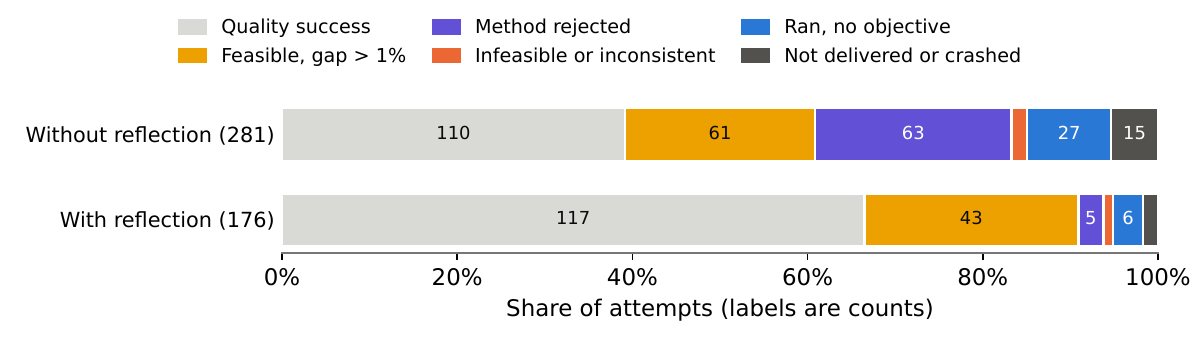}
\caption{Outcomes of retry attempts two to ten by the first check failed, split by policy. The shared first attempt is excluded, and each physical attempt is counted once. Segment labels give counts; the smallest segments are unlabeled.}
\label{fig:failure-retry}
\end{figure}

Reflection mainly speeds recovery, especially from failures whose feedback names the defect. Of the 119 pairs whose shared first attempt fails, a single reflection retry solves 94, compared with 67 for regeneration (Q@2 of 947 versus 920 of 972). By budget three, reflection solves 963 pairs, a level regeneration reaches only at budget ten. Regeneration catches up with more attempts, so the 27-pair advantage at budgets two and three narrows to seven at budget ten (970 versus 963). Part of this advantage depends on the status rule (\cref{sec:robustness}). \cref{fig:failure-retry} shows the advantage at the attempt level. Reflection turns 117 of its 176 retries (66\%) into successes, compared with 110 of 281 (39\%) under regeneration. The difference is largest for method rejections. Regeneration starts each retry from the original task, and the audit rejects 64 of its retries, 63 of them for an unproven optimality claim. Under reflection, the diagnosis step reads the judge's rationale and passes its advice to the retry. After a method rejection, the next reflection attempt passes the method audit in 42 of 44 cases, compared with 40 of 101 under regeneration (\cref{tab:after-rejection}), and the audit rejects only five reflection retries in total. For Claude Sonnet 5, reflection raises Q@3 from 61 to 78 of 81 instances. Reflection retries also receive the previous solver, and the protocol does not separate its effect from that of the rationale and the advice.

Objective gaps account for 43 of the 59 unsuccessful reflection retries, concentrated in FCTP (28) and two-stage stochastic planning (11; \cref{tab:retry-failures}). For these attempts, the feedback says only that the objective misses the target. The two pairs that remain unsolved after ten reflection attempts, GPT-6 Luna on a medium TSSO instance and Claude Sonnet 5 on a large FCTP instance, both end with this kind of failure. The benefit of reflection also varies across models. GPT-6 Luna, whose failures are mostly objective gaps, solves 75 of 81 instances within three attempts with reflection and 76 without it, and reaches higher coverage with reflection only at budget ten (80 versus 77). \cref{fig:reflection-gain} (\cref{app:retries}) reports the paired gains of every model at budgets two and three.

\FloatBarrier
\subsection{What does success cost?}
\label{sec:cost}

\begin{figure}[!htbp]
\centering
\includegraphics[width=.82\linewidth]{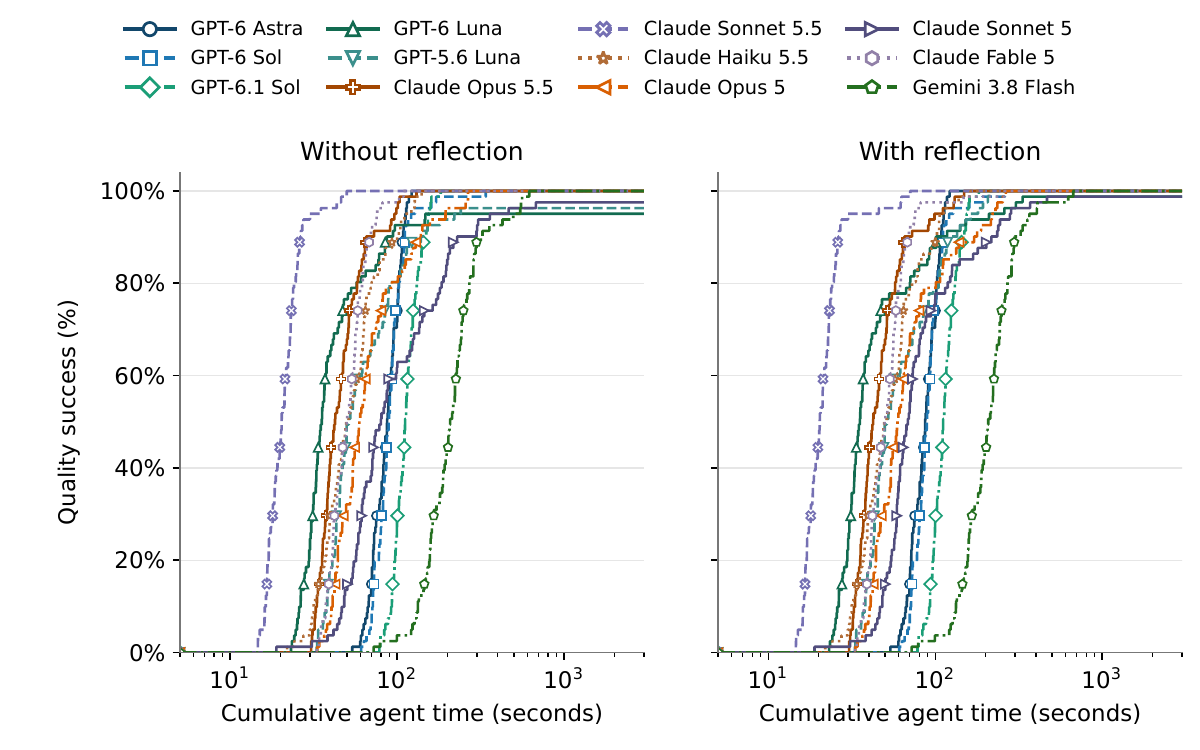}
\caption{Quality success within cumulative agent-time budgets, subject to the ten-attempt cap. A point counts the instances whose first quality success occurs within the indicated generation-plus-reflection time; all instances remain in the denominator. The time axis is logarithmic. Agent time excludes judge time, solver runs, and waiting imposed by scheduling or by provider usage limits.}
\label{fig:agent-time}
\end{figure}

Agent time separates models that tie on success. \cref{tab:resource-costs} reports the mean agent time and recorded tokens per instance through success or exhaustion, and \cref{fig:agent-time} counts the instances whose first quality success falls within each cumulative generation-plus-reflection time budget. Among models that solve every instance, Claude Sonnet 5.5 needs a mean of 21.5 seconds of agent time without reflection, GPT-6 Astra 87.3 seconds, GPT-6.1 Sol 114.8 seconds, and Gemini 3.8 Flash 228.2 seconds. Because an instance accumulates agent time until it is solved, faster recovery also saves time: reflection lowers the mean agent time of GPT-5.6 Luna from 82.5 to 70.8 seconds and of Claude Sonnet 5 from 121.4 to 107.9 seconds, despite the extra diagnosis steps.

\begin{table}[!htbp]
\centering
\caption{Mean cumulative agent time (seconds) and recorded context-plus-output tokens (thousands) per instance, through success or the ten-attempt limit. P and R denote regeneration and reflection. Means include all 81 instances. Token counts follow each tool's accounting (\cref{app:resources}); a dash marks Gemini 3.8 Flash, whose tool does not report a comparable total.}
\label{tab:resource-costs}
\small
\begin{tabular}{lrrrr}
\toprule
Model & Time P & Time R & Tokens P & Tokens R \\
\midrule
GPT-6 Astra & 87.3 & 87.3 & 85.7 & 85.7 \\
GPT-6 Sol & 93.8 & 93.1 & 139.2 & 141.2 \\
GPT-6.1 Sol & 114.8 & 114.8 & 92.2 & 92.2 \\
GPT-6 Luna & 56.7 & 60.5 & 199.5 & 254.3 \\
GPT-5.6 Luna & 82.5 & 70.8 & 277.6 & 283.1 \\
Claude Opus 5.5 & 48.8 & 49.9 & 75.9 & 77.7 \\
Claude Sonnet 5.5 & 21.5 & 22.4 & 60.9 & 63.3 \\
Claude Haiku 5.5 & 57.3 & 57.8 & 92.4 & 102.4 \\
Claude Opus 5 & 76.5 & 79.5 & 137.9 & 143.1 \\
Claude Sonnet 5 & 121.4 & 107.9 & 105.1 & 133.3 \\
Claude Fable 5 & 52.1 & 54.5 & 70.9 & 72.2 \\
Gemini 3.8 Flash & 228.2 & 223.0 & -- & -- \\
\bottomrule
\end{tabular}

\end{table}

\begin{figure}[!htbp]
\centering
\includegraphics[width=.82\linewidth]{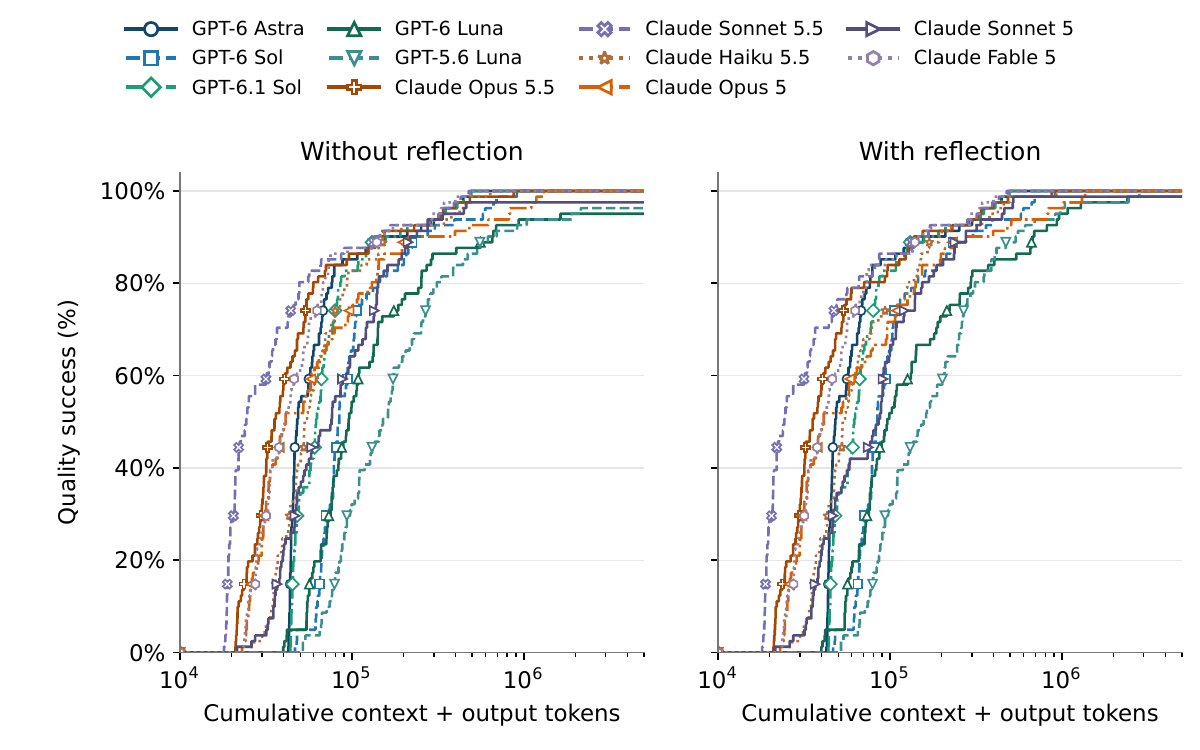}
\caption{Quality success within cumulative recorded-token budgets, without (left) and with (right) reflection, for the eleven models with comparable token totals. Each curve retains all 81 instances. The token axis is logarithmic; cached-input and reasoning tokens follow each provider's definitions (\cref{app:resources}).}
\label{fig:agent-tokens}
\end{figure}

Recorded tokens rank the models differently from time (\cref{fig:agent-tokens}). Over the eleven models with comparable token totals, the Spearman rank correlation between mean agent time and mean tokens without reflection is 0.47. The contrast is clearest among the five GPT models, which share the Codex tool and its token accounting. GPT-6 Luna and GPT-5.6 Luna finish in 56.7 and 82.5 seconds but consume 199.5 and 277.6 thousand tokens. GPT-6 Astra and GPT-6.1 Sol take 87.3 and 114.8 seconds and consume 85.7 and 92.2 thousand. The Luna models process 2.2 to 3.2 times as many tokens as GPT-6 Astra and GPT-6.1 Sol. Claude Sonnet 5.5 uses the least agent time and the fewest tokens, 60.9 thousand. Reflection raises token use for nine of the eleven models because each revision carries the previous solver and the diagnosis. Claude Sonnet 5, for example, consumes 133.3 thousand tokens with reflection and 105.1 thousand without, while its agent time falls. Token use is unchanged for GPT-6 Astra and GPT-6.1 Sol, which never retry. \cref{app:resources} describes the time attribution and the provider-specific token accounting.

\FloatBarrier
\subsection{Robustness of the findings}
\label{sec:robustness}

The unproven optimality rule determines Claude Sonnet 5's standing. \cref{tab:status-sensitivity} in \cref{app:unproven} reports every model's results with these rejections counted as passes when the output was otherwise feasible and within 1\% of the optimal objective. Claude Sonnet 5's Q@1 rises from 46 to 66 of 81, and no other model's Q@1 changes by more than one instance. Reflection's advantage over regeneration falls from 27 to 8 pairs at budget two (951 versus 943 of 972), from 27 to 10 at budget three (regeneration reaches 953), and from 7 to 5 at budget ten. Claude Sonnet 5 accounts for all 17 additional regeneration successes at budget three: its count rises from 61 to 78, equal to its reflection count.

First-attempt success varies little with instance size: 289 of the 324 pairs succeed on small instances, 283 on medium, and 281 on large (\cref{tab:size-results} in \cref{app:heterogeneity}).

\FloatBarrier

\FloatBarrier
\section{Conclusion}
\label{sec:conclusion}

On DecompGym, current coding agents design and implement both decomposition algorithms correctly in 853 of 972 first attempts (87.8\%). GPT-6 Astra and GPT-6.1 Sol write Benders and column-generation solvers that meet all three criteria on every instance at the first attempt, and seven of the twelve models succeed on more than 90\% of first attempts. The remaining failures depend on the method. Benders solvers most often return no finite objective or a suboptimal solution, and column-generation solvers most often fail the method audit, mainly by claiming optimality that their algorithm has not proved (92 of the 96 such claims come from Claude Sonnet 5). Most method-audit rejections (93 of 108, the method-conformance gap) return solutions that pass every output check, so rejecting them requires inspecting how the solver produced its answer. After a method-audit rejection, for which the judge's rationale names the defect, the next reflection attempt meets all three criteria in 41 of 44 cases. It repairs only 25 of 64 objective gaps, after which the feedback says only that the target is unmet. Time and token use separate models that reach the same final coverage.

The study uses a fixed synthetic inventory of moderate-size instances with a prescribed decomposition. The results show which models implement the methods correctly within a common budget. How the resulting solvers scale to larger instances is untested. Each model, instance, and policy contributes one saved trajectory, so paired differences are descriptive and additional repetitions would estimate run-to-run variation. A single GPT-6 Astra judge at high reasoning effort assesses method use from the source code, and GPT-6 Astra is also one of the evaluated models, run in separate sessions. LLM judges can favor outputs of their own model~\citep{panickssery2024llm}. We read every rejection rationale to categorize the findings but did not re-derive each verdict from the code. In addition, accepted attempts were not reviewed by hand, so the rate at which the judge accepts a solver that departs from the prescribed method is unmeasured. Agent time includes the speed of each tool and of its provider's service, and token totals follow provider-specific accounting. 

Agents built for decomposition will need checks on how their solvers reach an answer and feedback that locates the error, such as the bounds and gap that the solver reports at termination.

\FloatBarrier
\bibliographystyle{abbrvnat}
\bibliography{ref}
\appendix
\section*{Appendix}
\section{Problem Formulations and Reference Decompositions}
\label{app:formulations}

This appendix states, for each family, the original optimization model and the decomposition used by the reference implementation: the master problem, the subproblem or pricing problem, and the cut or column returned to the master. The agents never saw these formulations. Each agent received the natural-language task, the instance data, the output format, and the method contract in \cref{prompt:generation}, and designed its own models. The instance generators, reference implementations, stored optimal objectives, and details such as dual signs, initialization, and convergence traces are in the DecompGym repository at \url{https://github.com/SIKAI-C/DecompGym}.

The stored optimal objective of each instance is obtained by solving its original model, the full MIP stated below, to optimality with Gurobi. The quality criterion compares every agent's objective with this value.

Each reference implementation applies the prescribed decomposition under the agents' execution conditions: a 60-second limit on the whole solver process and one solver thread. It meets all three criteria on all 81 instances; the slowest run takes 27.25 seconds, and the largest deviation from the stored optimal objective is 0.14\%. A correct decomposition therefore fits comfortably within the time limit.

The reference implementations follow the same status rule as the agents: they report ``optimal'' only when their own bounds prove it. They report ``optimal'' on 54 instances and ``feasible'' on 27, and on 24 of those 27 the objective equals the stored optimal objective. A ``feasible'' label therefore means that the decomposition found a solution without certifying it. The 27 include 14 column-generation instances of BP and CSP, where the final integer restricted master does not reach the rounded-up linear bound; 8 TSSO instances, which stop at a 0.5\% gap between their bounds; 2 LRP instances, where the route-selection cuts give only a bound; and one instance each of UFL, KM, and FCTP, which stop at a 0.01\% target gap with the optimal objective.

Throughout, a bar denotes a value fixed by the master, a star an optimal dual solution, and a hat a dual ray.

\subsection{Benders families}
\label{app:form-benders}
In each Benders family, the master chooses binary structural decisions $y$ and an estimate $\theta$ of the remaining cost; a subproblem fixes $y=\bar y$, solves a linear program, and returns an optimality cut or, when infeasible, a feasibility cut. The master value is a lower bound and the best evaluated plan an upper bound; the reference stops when they meet within tolerance.

\paragraph{Uncapacitated facility location (UFL).}
Facilities $i$ have opening costs $f_i$, and serving customer $j$ from $i$ costs $D_{ij}$:
\[
\min_{y,x}\ \sum_i f_iy_i+\sum_{i,j}D_{ij}x_{ij}\quad\text{s.t.}\quad \sum_ix_{ij}=1\ \ \forall j,\qquad x_{ij}\le y_i\ \ \forall i,j,\qquad x\ge0,\ y\in\{0,1\}.
\]
The master minimizes $\sum_if_iy_i+\sum_j\theta_j$ subject to $\sum_iy_i\ge1$ and the cuts. For fixed openings, each customer solves $Q_j(\bar y)=\min\{\sum_iD_{ij}x_i:\sum_ix_i=1,\ 0\le x_i\le\bar y_i\}$, whose dual $(\alpha_j,\lambda_{ij}\le0)$ gives the disaggregated optimality cut $\theta_j\ge\alpha_j^\star+\sum_i\lambda_{ij}^\star y_i$. The constraint $\sum_iy_i\ge1$ keeps every subproblem feasible.

\paragraph{Fixed-charge transportation (FCTP).}
Suppliers $i$ have supply $s_i$, demand nodes $j$ have demand $d_j$, and arc $(i,j)$ has capacity $u_{ij}$, fixed charge $f_{ij}$, and unit cost $c_{ij}$:
\[
\min_{y,x}\ \sum_{i,j}f_{ij}y_{ij}+\sum_{i,j}c_{ij}x_{ij}\quad\text{s.t.}\quad \sum_jx_{ij}\le s_i\ \ \forall i,\qquad \sum_ix_{ij}\ge d_j\ \ \forall j,\qquad x_{ij}\le u_{ij}y_{ij},\qquad x\ge0,\ y\in\{0,1\}.
\]
The master chooses open arcs, with $\sum_iy_{ij}\ge1$ for every $j$, and an estimate $\theta$ of shipment cost. The subproblem is the transportation linear program over the open arcs. If it is feasible, its duals $(\pi\le0,\sigma\ge0,\mu\le0)$ give $\theta\ge\sum_is_i\pi_i^\star+\sum_jd_j\sigma_j^\star+\sum_{i,j}u_{ij}\mu_{ij}^\star y_{ij}$. If the open arcs cannot carry the demand, a dual ray gives the feasibility cut $\sum_is_i\hat\pi_i+\sum_jd_j\hat\sigma_j+\sum_{i,j}u_{ij}\hat\mu_{ij}y_{ij}\le0$, which removes the current plan. Feasibility cuts dominate the early iterations: on the large instance with seed 0, the reference needs 207 iterations to close its bounds.

\paragraph{$k$-median (KM).}
Exactly $k$ of the candidate locations $i$ must be selected, and customers $j$ are served at cost $D_{ij}$:
\[
\min_{y,x}\ \sum_{i,j}D_{ij}x_{ij}\quad\text{s.t.}\quad \sum_iy_i=k,\qquad \sum_ix_{ij}=1\ \ \forall j,\qquad 0\le x_{ij}\le y_i,\qquad y\in\{0,1\}.
\]
The master minimizes $\theta$ subject to $\sum_iy_i=k$ and the cuts. The subproblem solves the assignment linear program for the selected locations, and its duals give the aggregate optimality cut $\theta\ge\sum_j\alpha_j^\star+\sum_{i,j}\lambda_{ij}^\star y_i$. Any selection with $k\ge1$ admits an assignment, so no feasibility cuts arise.

\paragraph{Location routing (LRP).}
Depots $j$ have opening costs $f_j$ and fleet limits $M_j$, and a supplied set $R$ of feasible routes gives each route $r$ a depot $j(r)$, a cost $c_r$, and customer coverage $a_{ir}\in\{0,1\}$:
\[
\min_{y,\lambda}\ \sum_jf_jy_j+\sum_{r\in R}c_r\lambda_r\quad\text{s.t.}\quad \sum_ra_{ir}\lambda_r=1\ \ \forall i,\qquad \sum_{r:j(r)=j}\lambda_r\le M_jy_j\ \ \forall j,\qquad \sum_jy_j\ge1,\qquad y,\lambda\in\{0,1\}.
\]
The master chooses depots and a routing-cost estimate. The cut-generating subproblem is the linear relaxation of route selection for the open depots; its duals $(\alpha,\beta\le0)$ give $\theta\ge\sum_i\alpha_i^\star+\sum_jM_j\beta_j^\star y_j$, and a dual ray gives a feasibility cut of the same form. Because these cuts describe continuous recourse, the reference also solves the binary route selection at each master solution to obtain an integer incumbent. The bounds can therefore remain apart: on the small instance with seed 0, the reference stops with a lower bound of 15.541 and an integer cost of 15.745. The monolithic reference confirms that 15.745 is optimal, but the decomposition alone does not prove it, so the correct status is ``feasible''.

\paragraph{Two-stage stochastic planning (TSSO).}
Before demand is known, the planner activates regular capacity $K_{it}$ for product $i$ in period $t$ at cost $F_{it}$. In each scenario $s$ with probability $p_s$ and demand $d_{its}$, it then chooses regular production $r$, overtime $o$, inventory $I$, and backlog $B$:
\[
\begin{aligned}
\min\ & \sum_{i,t}F_{it}y_{it}+\sum_sp_s\sum_{i,t}\bigl(c^{\mathrm{reg}}_{it}r_{its}+c^{\mathrm{ot}}_{it}o_{its}+h_{it}I_{its}+b_{it}B_{its}\bigr)\\
\text{s.t.}\ & I_{i,t-1,s}-B_{i,t-1,s}+r_{its}+o_{its}=d_{its}+I_{its}-B_{its},\qquad r_{its}\le K_{it}y_{it},\qquad y\in\{0,1\},\ r,o,I,B\ge0.
\end{aligned}
\]
The master chooses one activation plan $y$ and an estimate $\theta$ of expected operating cost. Each scenario solves a production-planning linear program at $\bar y$; unlimited overtime and backlog keep it feasible. The capacity duals $\mu_{its}^\star\le0$ define $\alpha_s=Q_s(\bar y)-\sum_{i,t}\mu^\star_{its}K_{it}\bar y_{it}$, and the scenarios combine into one optimality cut $\theta\ge\sum_sp_s\bigl(\alpha_s+\sum_{i,t}\mu^\star_{its}K_{it}y_{it}\bigr)$. The reference stops at a 0.5\% gap between its own bounds and returns the best complete incumbent.

\subsection{Column-generation families}
\label{app:form-cg}
In each column-generation family, a restricted master linear program chooses weights for the columns generated so far, and a pricing problem uses its duals to find a column with negative reduced cost. After pricing finds none, the reference solves the master over the generated columns with integer variables. As explained in \cref{app:unproven}, that integer solution is proved optimal only when it attains the rounded lower bound from the completed linear relaxation.

\paragraph{Bin packing (BP).}
Item types $t$ have size $s_t$ and multiplicity $d_t$, and bins have capacity $C$. A pattern $p$ packs $a_{tp}$ items of type $t$ with $\sum_ts_ta_{tp}\le C$. The original problem minimizes the number of bins that cover all items, and the master is
\[
\min_{\lambda\ge0}\ \sum_{p\in P'}\lambda_p\quad\text{s.t.}\quad\sum_{p\in P'}a_{tp}\lambda_p\ge d_t\ \ \forall t,
\]
over the generated patterns $P'$, initialized with single-type patterns. With covering duals $\pi_t\ge0$, pricing is the integer knapsack $\max\{\sum_t\pi_ta_t:\sum_ts_ta_t\le C,\ a\in\mathbb{Z}^{T}_{\ge0}\}$, and a pattern whose value exceeds one enters the master. The reference also adds the alternative knapsack solutions that Gurobi finds during pricing as columns, then removes surplus bins from the integer solution before mapping types to item identifiers.

\paragraph{Cutting stock (CSP).}
Stock pieces have width $W$, and piece type $i$ has width $w_i$ and demand $d_i$. The master and pricing problems have the same form as for bin packing, with patterns satisfying $\sum_iw_ia_{ip}\le W$:
\[
\min_{\lambda\ge0}\ \sum_{p\in P'}\lambda_p\ \ \text{s.t.}\ \ \sum_{p\in P'}a_{ip}\lambda_p\ge d_i\ \ \forall i;\qquad \bar c(a)=1-\sum_i\pi_ia_i,\quad \sum_iw_ia_i\le W,\ a\in\mathbb{Z}_{\ge0}.
\]
The initial columns cut $\lfloor W/w_i\rfloor$ pieces of a single type. This is the family of the case study in \cref{sec:case-study}.

\paragraph{Generalized assignment (GAP).}
Each job $i$ must be assigned to one machine $j$, at cost $c_{ij}$ and resource use $a_{ij}$, within capacity $b_j$:
\[
\min_x\ \sum_{i,j}c_{ij}x_{ij}\quad\text{s.t.}\quad \sum_jx_{ij}=1\ \ \forall i,\qquad \sum_ia_{ij}x_{ij}\le b_j\ \ \forall j,\qquad x\in\{0,1\}.
\]
A column is a capacity-feasible job subset $z_{\cdot jp}$ for one machine, with cost $c_{jp}=\sum_ic_{ij}z_{ijp}$. The master chooses one convex combination of patterns per machine:
\[
\min_{\lambda\ge0}\ \sum_j\sum_{p\in P'_j}c_{jp}\lambda_{jp}\quad\text{s.t.}\quad \sum_j\sum_{p\in P'_j}z_{ijp}\lambda_{jp}=1\ \ \forall i,\qquad \sum_{p\in P'_j}\lambda_{jp}=1\ \ \forall j.
\]
With job duals $\pi_i$ and convexity duals $\sigma_j$, pricing for machine $j$ is the binary knapsack $\min\{\sum_i(c_{ij}-\pi_i)z_i-\sigma_j:\sum_ia_{ij}z_i\le b_j,\ z\in\{0,1\}\}$. The reference starts from empty and single-job patterns and reaches a feasible master through a phase-I with artificial variables and the same pricing, without solving a full assignment model.

\paragraph{Shift scheduling (SHIFT).}
Time slot $t$ requires $d_t$ workers, and a legal shift $p=[s_p,e_p]$ is a contiguous interval whose length lies between $L_{\min}$ and $L_{\max}$; $a_{tp}=1$ when the shift covers slot $t$. The original model and its master are
\[
\min_x\ \sum_{p\in P}x_p\quad\text{s.t.}\quad\sum_{p\in P}a_{tp}x_p\ge d_t\ \ \forall t,\quad x\in\mathbb{Z}_{\ge0};\qquad \min_{\lambda\ge0}\ \sum_{p\in P'}\lambda_p\quad\text{s.t.}\quad\sum_{p\in P'}a_{tp}\lambda_p\ge d_t\ \ \forall t,
\]
where the master starts from all minimum-length shifts. Pricing finds a legal interval with $1-\sum_{t=s}^{e}\pi_t<0$; the reference enumerates the legal intervals. On the small instance with seed 0, the final integer master attains the linear bound of 10 worker-shifts, which proves its optimality.

\FloatBarrier
\section{Instance Inventory}
\label{app:inventory}

\cref{tab:instance-sizes} reports the smallest and largest dimensions among the three instances of each family and size, generated with seeds 0, 1, and 2.

\begin{table}[!htbp]
\centering
\caption{Realized dimensions of the fixed instances. Each entry gives the range over the three seeds, in the order listed in the dimensions column.}
\label{tab:instance-sizes}
\small
\begin{tabular}{lllll}
\toprule
Family & Dimensions & Small & Medium & Large \\
\midrule
UFL & Sites / customers & 7--10 / 19--25 & 15--19 / 38--49 & 29--37 / 75--98 \\
FCTP & Origins / destinations & 2--3 / 3--4 & 3--4 / 4--5 & 4--5 / 6--7 \\
KM & Sites / customers / $k$ & 7--10 / 12--17 / 2 & 12--15 / 22--27 / 4 & 17--20 / 32--37 / 4--6 \\
LRP & Depots / customers & 3--4 / 6--7 & 5--6 / 8--9 & 6--7 / 11--13 \\
TSSO & Products / periods / scenarios & 2--3 / 3--4 / 2--3 & 4--5 / 6--7 / 3--5 & 6--7 / 8--9 / 6--8 \\
BP & Items / bin capacity & 8--14 / 72--117 & 14--25 / 87--152 & 18--29 / 107--168 \\
CSP & Types / stock length & 2--3 / 131--192 & 3--6 / 161--247 & 4--7 / 191--277 \\
GAP & Machines / jobs & 2 / 4--5 & 2--3 / 5--8 & 2--3 / 6--9 \\
SHIFT & Time slots / max. shift & 6--7 / 5 & 8--11 / 5 & 9--12 / 5 \\
\bottomrule
\end{tabular}

\end{table}

\FloatBarrier
\section{Scoring Definitions}
\label{app:scoring}

For attempt $t$ from model $m$ on instance $i$ under policy $p$, let $E_{mit}^{p}$ indicate successful execution, and let $V_{mit}^{p}$ indicate that execution succeeds, the output passes the numerical checks, and the method judge returns a pass. With recomputed objective $z_{mit}^{p}$ and stored optimal objective $z_i^{\mathrm{ref}}$, quality success is
\begin{equation}
Q_{mit}^{p}=V_{mit}^{p}\,\mathbf{1}\!\left\{\frac{|z_{mit}^{p}-z_i^{\mathrm{ref}}|}{\max(1,|z_i^{\mathrm{ref}}|)}\leq 0.01\right\},
\label{eq:quality}
\end{equation}
so $Q_{mit}^{p}\leq V_{mit}^{p}\leq E_{mit}^{p}$. The stored optimal objective $z_i^{\mathrm{ref}}$ is obtained by solving the full MIP of instance $i$ to optimality with Gurobi. Missing or nonfinite objectives fail the numerical checks and therefore both V and Q. For $G\in\{E,V,Q\}$, cumulative success at budget $k$ is $G_{mi}^{p}(k)=\max_{t\leq k}G_{mit}^{p}$, so an early success remains a success at later budgets without further agent runs.

Each of the 972 model--instance pairs is run under both policies, giving 1,944 policy trajectories. The shared first attempt counts toward both policies when they are compared and once in totals of agent runs, so the failure analysis in \cref{sec:failures} covers 1,429 physical attempts: 972 first attempts and 457 retries. Size comparisons use 324 pairs per size level, and reflection gains are paired differences on the same instances. Success within a time or token budget follows the same cumulative rule.

\FloatBarrier
\section{Execution, Verification, and Quality}
\label{app:stages}

\begin{figure}[!htbp]
\centering
\includegraphics[width=.66\linewidth]{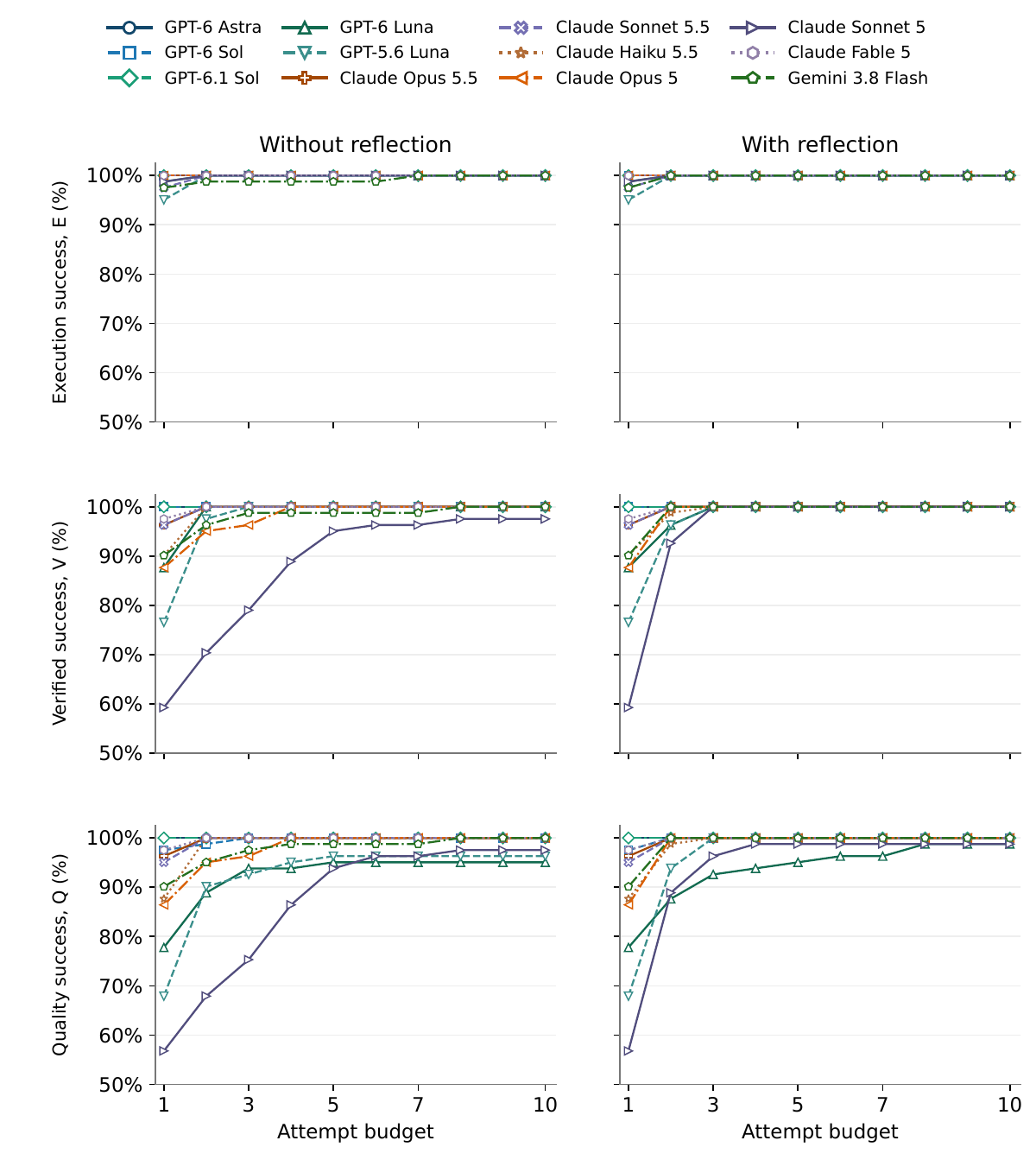}
\caption{Execution (E), verified feasibility and method use (V), and verified objective quality (Q) over attempt budgets. Each row uses the same vertical scale starting at 50\%, and each model uses the same 81 instances throughout.}
\label{fig:stages}
\end{figure}

\cref{tab:staged-counts} reports pooled cumulative successes under each criterion, and \cref{fig:stages} shows the cumulative E, V, and Q rates of each model. The drop from execution to quality is largest at the first attempt: 962 of 972 first attempts run, 876 also pass verification, and 853 meet all three criteria. Retries close the execution gap first. With reflection, every pair passes execution by budget two and verification by budget three, when nine pairs still miss the quality target; two remain after ten attempts.

\begin{table}[!htbp]
\centering
\caption{Cumulative successes by criterion among 972 complete model--instance pairs. P denotes independent regeneration and R reflection. Criterion V includes numerical feasibility and the method audit.}
\label{tab:staged-counts}
\small
\begin{tabular}{lrrrrrrr}
\toprule
Criterion & @1 & @2 P & @2 R & @3 P & @3 R & @10 P & @10 R \\
\midrule
E: executable & 962 & 971 & 972 & 971 & 972 & 972 & 972 \\
V: verified & 876 & 939 & 959 & 951 & 972 & 970 & 972 \\
Q: quality & 853 & 920 & 947 & 936 & 963 & 963 & 970 \\
\bottomrule
\end{tabular}

\end{table}

\FloatBarrier
\section{Method-Audit Findings}
\label{app:judge}

The method audit rejects 108 of the 1,429 physical attempts; \cref{prompt:judge} shows the judge's instructions. Each rejection rationale was read by hand and assigned the one finding it names as the failure; when a rationale lists several issues, the label follows the one it identifies as decisive. The labels and the decisive excerpt of each rationale are released with the analysis code.

\emph{Unproven optimality claim} (96 rejections). The solver reports its solution as ``optimal'' on a path where its algorithm has not proved global optimality. Typical cases map an optimal Gurobi status of the final integer restricted master directly to ``optimal'', or stop the pricing loop at an iteration or time limit and still report optimality. Ninety-five of these rejections concern column-generation solvers and one a Benders solver. Claude Sonnet 5 accounts for 92; GPT-6 Luna, Claude Haiku 5.5, and Claude Opus 5 account for the other four (\cref{tab:unproven-claims}). \cref{app:unproven} explains the rule and why it is part of the method contract.

\emph{Inactive decomposition} (8 rejections). The decomposition is written but does not operate. In six Benders solvers, the convergence test compares an infinite initial upper bound with itself, evaluates $\infty\leq\infty$ as true after the first master solve, and terminates before any subproblem is solved. Other cases include a master whose initial columns are never attached to its constraints, so the first solve is infeasible and pricing never runs, and a Benders master without the recourse variable in its objective, so cuts cannot influence the search.

\emph{Solution bypass} (4 rejections). A reachable fallback, such as first-fit-decreasing packing or enumeration of every admissible shift, produces the reported solution outside the decomposition.

\begin{table}[!htbp]
\centering
\caption{Unproven optimality claims by model and attempt type. First attempts are shared by both policies and counted once; P and R denote retries under regeneration and reflection. All 26 of Claude Sonnet 5's first-attempt claims are on column-generation instances, where 30 of its 36 first attempts fail (\cref{fig:family-first}).}
\label{tab:unproven-claims}
\small
\begin{tabular}{lrrrr}
\toprule
Model & First attempt & Retries P & Retries R & Total \\
\midrule
Claude Sonnet 5 & 26 & 62 & 4 & 92 \\
Claude Opus 5 & 1 & 1 & 0 & 2 \\
GPT-6 Luna & 1 & 0 & 0 & 1 \\
Claude Haiku 5.5 & 1 & 0 & 0 & 1 \\
Other eight models & 0 & 0 & 0 & 0 \\
\midrule
All models & 29 & 63 & 4 & 96 \\
\bottomrule
\end{tabular}

\end{table}

\FloatBarrier
\subsection{Unproven optimality claims}
\label{app:unproven}

Every solver reports a status label for its solution: ``optimal'', ``feasible'', ``infeasible'', or ``error''. The label ``optimal'' asserts that no better solution exists. The method contract, given identically to every agent and to the judge, ends with the sentence ``Use status feasible when returning an incumbent without a global optimality proof; optimality of an integer restricted master alone does not prove global optimality'' (highlighted in \cref{prompt:generation,prompt:judge}). The quality criterion checks the objective value and not the status label, so a solver that reports ``feasible'' can meet all three criteria: 115 of the 1,080 attempts that met all three criteria, including 70 column-generation solvers, reported ``feasible'' (\cref{tab:success-status}). The reference implementations themselves report ``feasible'' on 27 of the 81 instances (\cref{app:formulations}). An unproven optimality claim is rejected because the algorithm has not established it.

\begin{table}[!htbp]
\centering
\caption{Status reported by the 1,080 attempts that met all three criteria, by decomposition method. Each attempt is counted once, including the shared first attempt.}
\label{tab:success-status}
\small
\begin{tabular}{lrrr}
\toprule
Reported status & Benders & Column generation & All \\
\midrule
\code{optimal} & 548 & 417 & 965 \\
\code{feasible} & 45 & 70 & 115 \\
\midrule
All successful attempts & 593 & 487 & 1{,}080 \\
\bottomrule
\end{tabular}

\end{table}

Consider cutting stock, a minimization problem. Let $\Omega$ be the set of all feasible cutting patterns, $K\subseteq\Omega$ the patterns generated so far, and $z^\star$ the optimal number of stock pieces. Column generation solves the linear relaxation of the master over $K$ and prices new patterns with its duals. Two bounds apply.
\begin{itemize}[leftmargin=1.5em,itemsep=1pt,topsep=2pt]
\item If pricing proves that no pattern in $\Omega$ has negative reduced cost, the restricted linear value equals the full linear relaxation value $z_{\mathrm{LP}}$, and $z_{\mathrm{LP}}\leq z^\star$. This is a valid lower bound. If pricing stopped early at an iteration or time limit, the restricted value does not certify a lower bound.
\item Solving the restricted master as an integer program over $K$ gives a feasible solution with value $z_{\mathrm{IP}}(K)\geq z^\star$. This is an upper bound. Gurobi's ``optimal'' status for this solve means only that no better integer solution exists \emph{using the patterns in $K$}. Solving the integer restricted master once after the linear relaxation has converged is the heuristic known as price-and-branch~\citep{lubbecke2005selected}.
\end{itemize}
Hence $z_{\mathrm{LP}}\leq z^\star\leq z_{\mathrm{IP}}(K)$. Because the objective is integer, the solution is proved optimal when $z_{\mathrm{IP}}(K)=\lceil z_{\mathrm{LP}}\rceil$. Otherwise the gap is unresolved: the optimal integer solution may need a pattern that pricing never generated, because pricing selects patterns that improve the linear relaxation and does not consider the integer program. Branch-and-price closes such a gap by continuing to generate columns inside branch-and-bound~\citep{barnhart1998branch,lubbecke2005selected}. A solver that reports the integer master's optimal status as ``optimal'' for the whole problem conflates optimality over $K$ with optimality over $\Omega$. The same reasoning applies to the other column-generation families, and an analogous argument, in which the Benders master supplies the lower bound, applies to the single Benders case. \cref{sec:case-study} applies the argument to a cutting-stock instance.

A reported optimality status is a correctness claim that users act on, for example by stopping a search or trusting a bound, so an unproven optimality claim is an error even when the objective happens to be optimal. Because the rule decides one model's standing, \cref{tab:status-sensitivity} reports every model's success counts with these rejections counted as passes; \cref{sec:robustness} summarizes the changes, and the reflection counts at budgets three and ten are unchanged.

\begin{table}[!htbp]
\centering
\caption{Solved instances (of 81) under the status rule used in the paper and, after the arrow where different, with unproven optimality claims counted as passes when the output was otherwise feasible and within 1\% of the optimal objective. A trajectory stops at its first success, so the relaxed counts follow each saved trajectory up to its first accepted attempt. P denotes independent regeneration and R reflection.}
\label{tab:status-sensitivity}
\small
\begin{tabular}{lccccccc}
\toprule
Model & Q@1 & Q@2 P & Q@2 R & Q@3 P & Q@3 R & Q@10 P & Q@10 R \\
\midrule
GPT-6 Astra & 81 & 81 & 81 & 81 & 81 & 81 & 81 \\
GPT-6 Sol & 79 & 80 & 81 & 81 & 81 & 81 & 81 \\
GPT-6.1 Sol & 81 & 81 & 81 & 81 & 81 & 81 & 81 \\
GPT-6 Luna & 63 & 72 & 71 & 76 & 75 & 77 & 80 \\
GPT-5.6 Luna & 55 & 73 & 76 & 75 & 81 & 78 & 81 \\
Claude Opus 5.5 & 78 & 81 & 81 & 81 & 81 & 81 & 81 \\
Claude Sonnet 5.5 & 77 & 81 & 81 & 81 & 81 & 81 & 81 \\
Claude Haiku 5.5 & 71 $\to$ \textbf{72} & 81 & 80 & 81 & 81 & 81 & 81 \\
Claude Opus 5 & 70 $\to$ \textbf{71} & 77 $\to$ \textbf{78} & 81 & 78 & 81 & 81 & 81 \\
Claude Sonnet 5 & 46 $\to$ \textbf{66} & 55 $\to$ \textbf{77} & 72 $\to$ \textbf{76} & 61 $\to$ \textbf{78} & 78 & 79 $\to$ \textbf{81} & 80 \\
Claude Fable 5 & 79 & 81 & 81 & 81 & 81 & 81 & 81 \\
Gemini 3.8 Flash & 73 & 77 & 81 & 79 & 81 & 81 & 81 \\
\midrule
All models (of 972) & 853 $\to$ \textbf{875} & 920 $\to$ \textbf{943} & 947 $\to$ \textbf{951} & 936 $\to$ \textbf{953} & 963 & 963 $\to$ \textbf{965} & 970 \\
\bottomrule
\end{tabular}

\end{table}

\FloatBarrier
\section{Problem Families and Instance Size}
\label{app:heterogeneity}

\begin{figure}[!htbp]
\centering
\includegraphics[width=.86\linewidth]{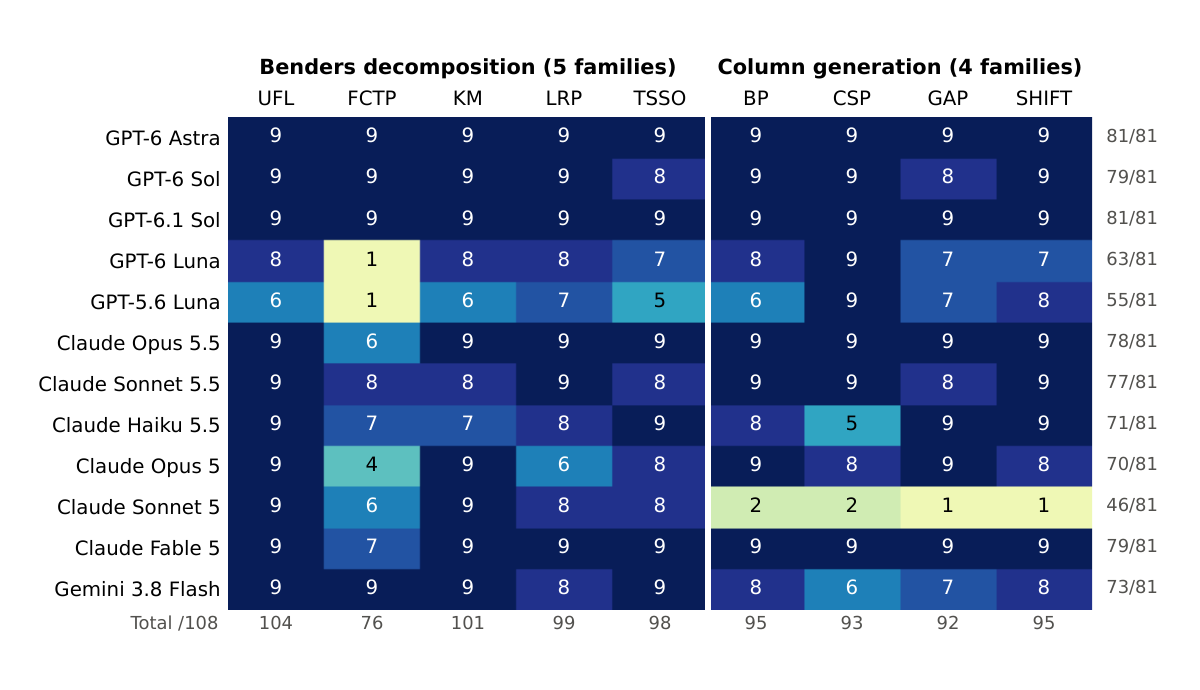}
\caption{First-attempt quality successes by model and family, out of nine instances per cell (three sizes $\times$ three seeds). The bottom row totals each family over the twelve models (out of 108), and the right column totals each model (out of 81).}
\label{fig:family-first}
\end{figure}

\cref{fig:family-first} reports first-attempt success for every model and family; \cref{sec:failures} discusses its FCTP column and its Claude Sonnet 5 row.

\FloatBarrier

\begin{figure}[!htbp]
\centering
\includegraphics[width=.86\linewidth]{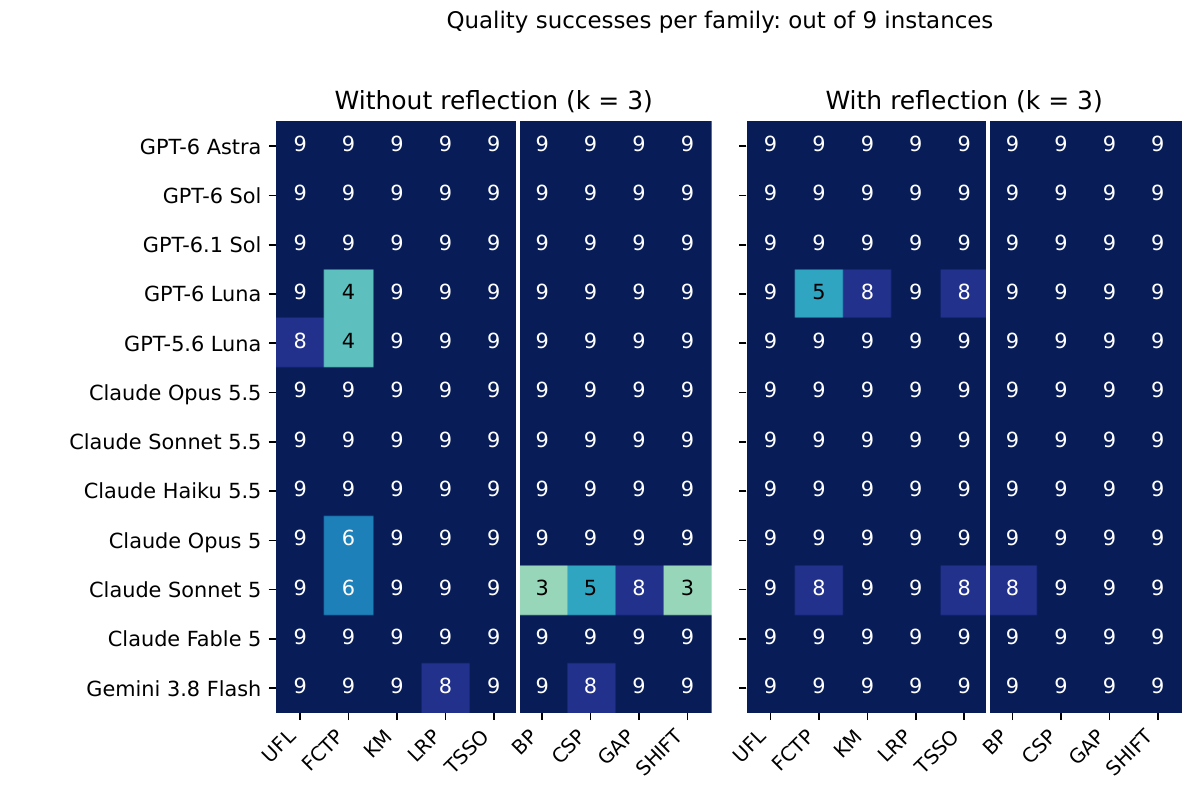}
\caption{Quality successes at budget three, out of nine instances per cell. One instance is 11.1 percentage points. Columns to the left of the white divider are Benders families; columns to its right are column-generation families.}
\label{fig:family-quality}
\end{figure}

\cref{fig:family-quality} repeats the matrix at budget three for both policies. Of the 27 net additional successes under reflection at that budget, Benders families account for 10 and column-generation families for 17. FCTP alone gains 11; UFL and LRP gain one each, and KM and TSSO lose one and two. Claude Sonnet 5's gain comes mostly from SHIFT (6), BP (5), and CSP (4).

\FloatBarrier

\begin{figure}[!htbp]
\centering
\includegraphics[width=\linewidth]{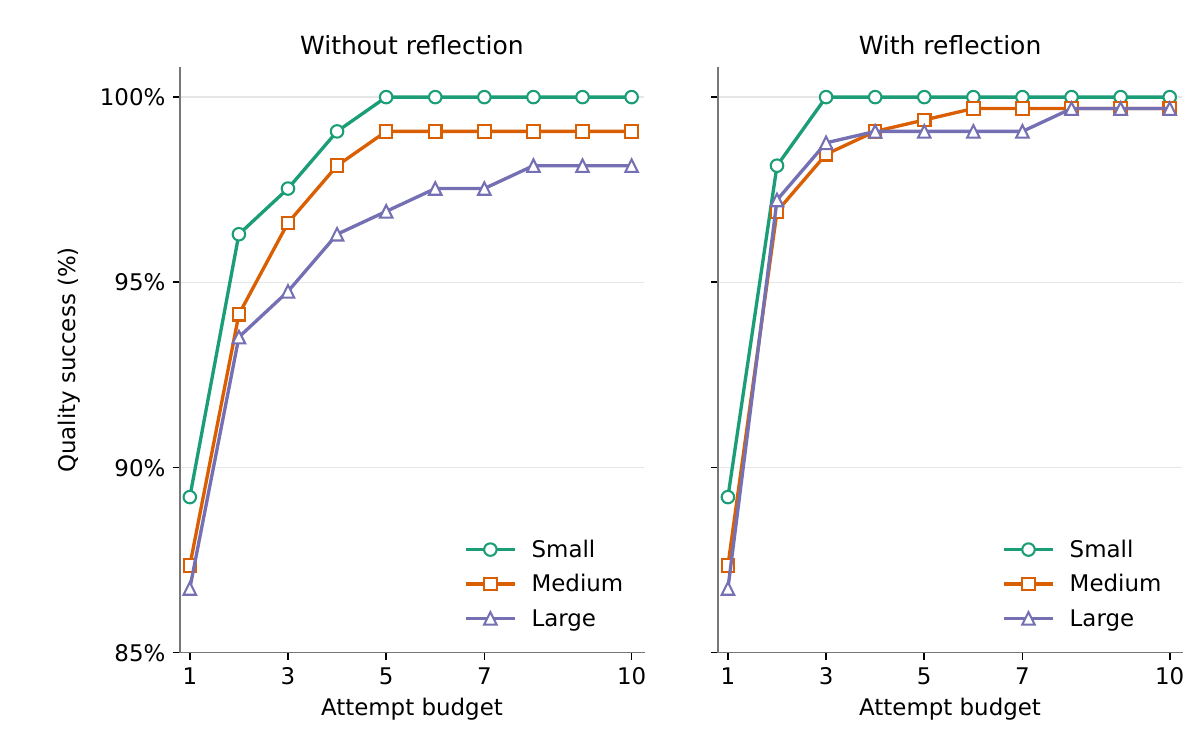}
\caption{Quality success by size over all attempt budgets. Each size contains three fixed instances from each of nine families, evaluated with all twelve models. The panels share axes, with the vertical scale starting at 85\%, and differ only in the retry policy.}
\label{fig:size-quality}
\end{figure}

Each size level contributes 324 model--instance pairs. First-attempt successes are 289 for small, 283 for medium, and 281 for large instances. At budget three, reflection adds 8, 6, and 13 successes over regeneration, respectively. \cref{tab:size-results} and \cref{fig:size-quality} give the full comparison.

\begin{table}[!htbp]
\centering
\caption{Quality success by size, with 324 pairs in each row. P denotes independent regeneration and R reflection.}
\label{tab:size-results}
\small
\begin{tabular}{lrrrrr}
\toprule
Size & Q@1 & Q@3 P & Q@3 R & Q@10 P & Q@10 R \\
\midrule
Small & 289 & 316 & 324 & 324 & 324 \\
Medium & 283 & 313 & 319 & 321 & 323 \\
Large & 281 & 307 & 320 & 318 & 323 \\
\bottomrule
\end{tabular}

\end{table}

\FloatBarrier
\section{Retries and Reflection}
\label{app:retries}

\begin{figure}[!htbp]
\centering
\includegraphics[width=.98\linewidth]{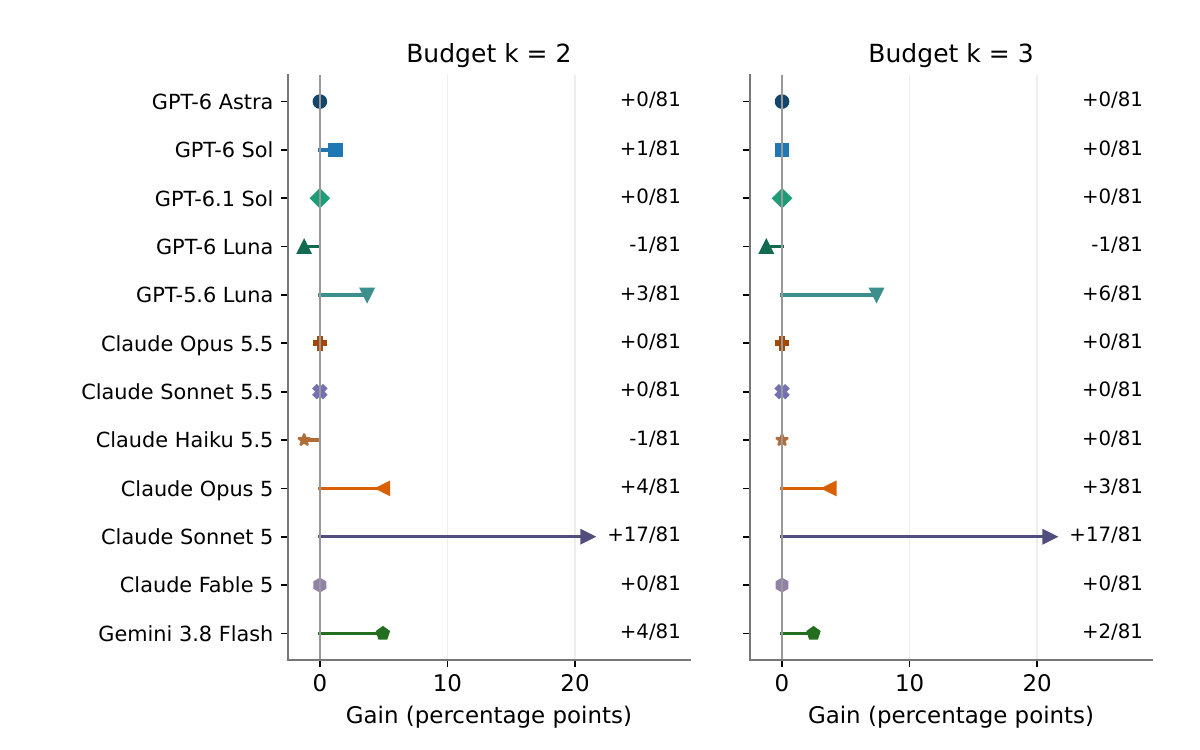}
\caption{Paired quality-success differences between reflection and regeneration at budgets two and three. Positive values favor reflection. Labels give the net number of additional instances solved out of 81. Each difference compares one saved trajectory per policy and instance.}
\label{fig:reflection-gain}
\end{figure}

\cref{fig:reflection-gain} breaks the 27-pair advantage of reflection at budgets two and three (\cref{sec:reflection}) down by model. Claude Sonnet 5 contributes 17 of the 27 at both budgets. GPT-5.6 Luna gains three instances at budget two and six at budget three, and Claude Opus 5 and Gemini 3.8 Flash gain between two and four. GPT-6 Luna and Claude Haiku 5.5 each show a one-instance negative difference at budget two, and GPT-6 Luna also at budget three.

\cref{tab:after-rejection} follows each method rejection to the next attempt of the same policy. Rejections at the tenth attempt have no next attempt and are omitted, which leaves 101 of the 103 rejections under regeneration and all 44 under reflection. The counts include rejections of the shared first attempt, which both policies follow.

\begin{table}[!htbp]
\centering
\caption{Outcome of the attempt that follows a method-audit rejection in the same trajectory. P and R denote regeneration and reflection.}
\label{tab:after-rejection}
\small
\begin{tabular}{lrr}
\toprule
 & P & R \\
\midrule
Method rejections followed by another attempt & 101 & 44 \\
Next attempt passes the method audit & 40 & 42 \\
Next attempt meets all three criteria & 38 & 41 \\
\bottomrule
\end{tabular}

\end{table}

\cref{tab:retry-failures} classifies the unsuccessful retries by the first check they fail, using the categories of \cref{fig:failure-retry}, and splits objective gaps by problem family. Under regeneration, FCTP accounts for 54 of the 61 objective gaps; under reflection, FCTP and TSSO together account for 39 of 43.

\begin{table}[!htbp]
\centering
\caption{Unsuccessful retries (attempts two to ten) by the first check failed, with objective gaps split by family. Other families are UFL, CSP, and GAP under regeneration and BP and KM under reflection. One regeneration retry that the judge rejects also fails a numerical check and is counted there, so the method-rejected count is 63 rather than the 64 in \cref{tab:judge-findings}. P and R denote regeneration and reflection.}
\label{tab:retry-failures}
\small
\begin{tabular}{lrr}
\toprule
First check failed & P & R \\
\midrule
Feasible, gap $>$ 1\% & 61 & 43 \\
\quad FCTP & 54 & 28 \\
\quad TSSO & 2 & 11 \\
\quad Other families & 5 & 4 \\
Method rejected & 63 & 5 \\
Ran, no objective & 27 & 6 \\
Infeasible or inconsistent & 5 & 2 \\
Not delivered or crashed & 15 & 3 \\
\midrule
All unsuccessful retries & 171 & 59 \\
\bottomrule
\end{tabular}

\end{table}

\FloatBarrier
\section{Time and Token Accounting}
\label{app:resources}

\begin{figure}[!htbp]
\centering
\includegraphics[width=\linewidth]{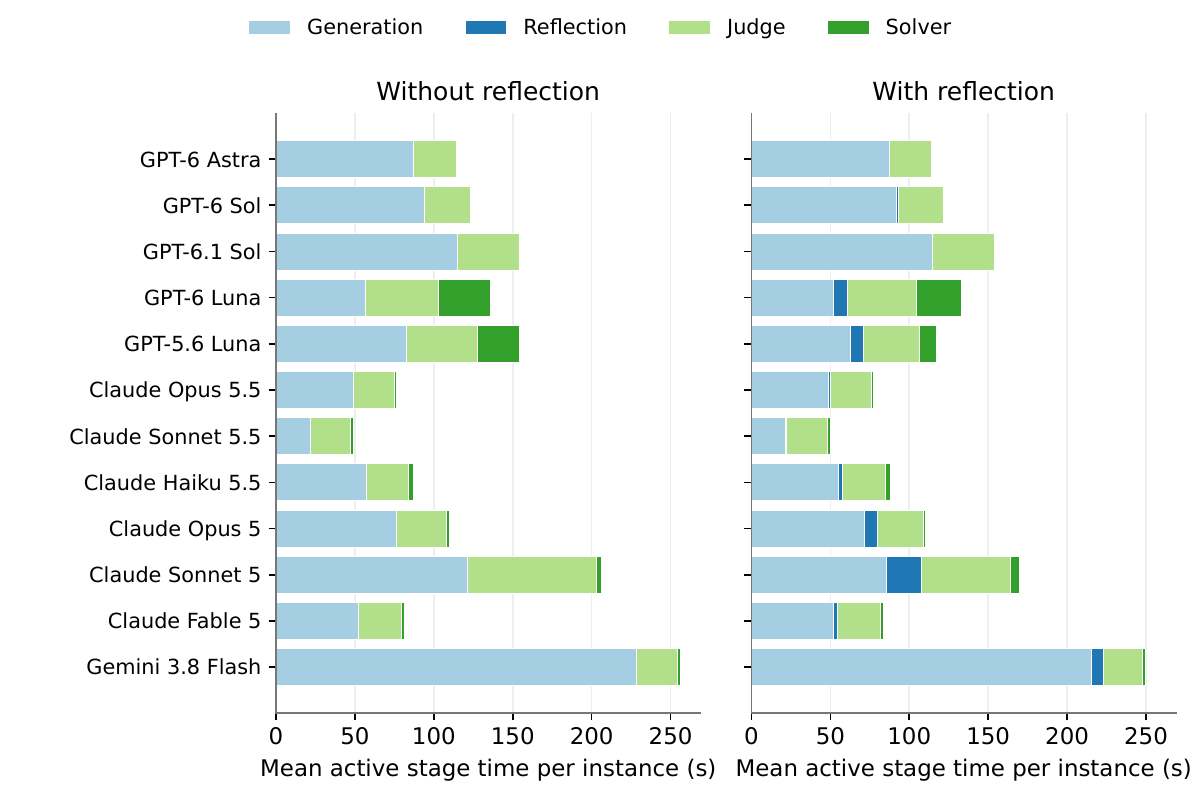}
\caption{Mean recorded active time per instance, decomposed into solver generation, reflection, method audit, and solver execution. The two panels share the time scale. All segments use recorded stage durations; they exclude waiting and scheduling intervals.}
\label{fig:cost-breakdown}
\end{figure}

Agent time sums the generation and reflection calls of each trajectory up to its first success or its tenth attempt, so \cref{tab:resource-costs} averages over all 81 instances, including trajectories that exhaust the budget. A reflection after attempt $t$ counts toward the cost of attempt $t+1$, and attempts after a success add nothing. \cref{fig:cost-breakdown} adds the time of the method audit and of solver execution. All times are recorded stage durations. They exclude user pauses and waits for scheduling or provider usage limits. Because stages ran in parallel, their sum differs from the wall-clock duration of the experiment. Calls that failed without producing an attempt remain in the raw call records and are excluded from these totals.

Recorded tokens follow each tool's accounting. For Codex, the total is input plus output tokens, and cached input is already part of input. For Claude Code, it is the sum of ordinary input tokens, tokens written to and read from the prompt cache, and output tokens. Reasoning tokens are part of output and are not added again. Counts accumulate over the agent's turns, so they measure the total volume processed over all turns; the providers also use different tokenizers. \cref{tab:resource-costs} and \cref{fig:agent-tokens} therefore report tokens for the eleven models with comparable totals.

Gemini's tool records output, thinking, cache-read, and its own total tokens, but not how cache reads relate to input, so no comparable total can be formed. \cref{tab:gemini-usage} reports the mean recorded output, thinking, cache-read, and provider-total tokens. Gemini 3.8 Flash remains in every attempt and time comparison.

\begin{table}[!htbp]
\centering
\caption{Gemini 3.8 Flash mean raw agent usage per instance, in thousands of tokens. Fields may overlap and must not be summed. The provider total is reproduced directly from the agent tool.}
\label{tab:gemini-usage}
\small
\begin{tabular}{lrrrr}
\toprule
Policy & Output & Thinking & Cache read & Provider total \\
\midrule
Plain & 33.6 & 30.1 & 99.7 & 129.0 \\
Reflection & 32.9 & 29.6 & 98.0 & 127.7 \\
\bottomrule
\end{tabular}

\end{table}

\FloatBarrier
\section{Configuration and Reproducibility}
\label{app:configuration}

The experiment uses Codex 0.162.1, Claude Code 2.1.296, and Google Antigravity 1.3.1. The generated solver environment uses Python 3.12.11, Gurobi 11.0.3, and macOS 26.4.1. The attempt limit is ten per policy, each agent call is limited to 900 seconds, and each solver run is limited to 60 seconds with one numerical solver thread. Each generation, reflection, and judge call runs in a new Docker container limited to one CPU and 1 GB of memory, with a newly created workspace. The agent tools run without persistent state: Codex with ephemeral sessions and user configuration and rules ignored; Claude Code without session persistence, MCP servers, or slash commands, and with tools limited to reading and editing files; Antigravity in a new project with its sandbox enabled. Generated solvers run outside the containers, on the host under the macOS sandbox. The machine runs two instances at a time, with at most two simultaneous agent calls for Codex and one each for Claude and Gemini. All generation and reflection calls use medium reasoning effort, and the method judge requests GPT-6 Astra at high effort. Model labels identify the requested models; the Antigravity tool does not independently report which model served a call.

The results combine two runs with the same instances, prompts, scoring code, and resource limits. The first contains the five GPT models, Gemini 3.8 Flash, Claude Opus 5.5, and Claude Sonnet 5.5; the second adds Claude Haiku 5.5, Claude Opus 5, Claude Sonnet 5, and Claude Fable 5. The released records identify the protocol of each run and contain all 972 complete model--instance pairs of the twelve models in \cref{tab:main-results}. \cref{app:formulations} reports the validation of the reference implementations.

The released analysis gives, for each instance, the first successful attempt, the success counts at every attempt budget, and the time and token totals. Every plotted count is checked against the paired outcomes, including the nesting $Q\leq V\leq E$ and the equality of the shared first-attempt scores across policies. Each model keeps one color, marker, and line style across figures, and panels that compare the two policies place regeneration on the left and reflection on the right, on matched axes. An index links each figure to its source data.

\FloatBarrier
\section{Prompt Examples}
\label{app:prompts}

The prompts below come from the case study in \cref{sec:case-study}: Claude Sonnet 5, through Claude Code, on the medium cutting-stock instance with seed 0. The judge prompt is the one GPT-6 Astra received for that attempt. Codex receives identical text. The Antigravity tool receives the same task and context with its workspace rules merged into one line, which also bars the agent from saved memory, add-on skills, helper agents, and earlier conversations. Every prompt is verbatim except where \textcolor{black!50}{gray} text marks an elided context block. The sentence printed in \textcolor{promptmark}{\textbf{red}} is the status requirement discussed in \cref{app:unproven}; the colors are added for the reader, and the agents received plain text. \cref{tab:prompt-calls} summarizes what each call receives.

\begin{table}[!htbp]
\centering
\caption{Agent and judge calls, their inputs, and what each call produces. Every call begins with the workspace rules shown at the top of \cref{prompt:generation}.}
\label{tab:prompt-calls}
\small
\begin{tabularx}{\linewidth}{>{\raggedright\arraybackslash}p{3.1cm}>{\raggedright\arraybackslash}X>{\raggedright\arraybackslash}p{2.3cm}l}
\toprule
Call & Inputs & Output & Prompt \\
\midrule
Generation & Task description, output format, method contract, run-time limits, instance data & Solver code & \cref{prompt:generation} \\
Method judge & Method contract and judging instructions, candidate solver code & Accept or reject decision with rationale & \cref{prompt:judge} \\
Reflection diagnosis & Full task prompt, instance data, failed solver code, execution output, judge rationale, returned solution, and, after an objective miss, a one-line quality message & Written advice & \cref{prompt:reflection} \\
Revision generation & Generation prompt, previous solver code, and the advice & Revised solver code & \cref{prompt:revision} \\
\bottomrule
\end{tabularx}
\end{table}

\subsection{Generation}
\label{app:prompt-generation}
\cref{prompt:generation} is the full first-attempt prompt, with the instance data abbreviated. Regeneration retries receive the same text in a fresh call. The task description and output format are family-specific; the method contract consists of one method-specific paragraph followed by three paragraphs shared by all nine families. \cref{prompt:benders} shows the method-specific paragraph that Benders families receive in its place.

\promptfile{Generation prompt (cutting stock, column generation)}{generation_csp.tex}{prompt:generation}

\promptfile{Method-specific contract paragraph for Benders families (shown for UFL)}{contract_benders.tex}{prompt:benders}

\subsection{Method judge}
\label{app:prompt-judge}
The judge, GPT-6 Astra at high effort, receives the contract that the agent received and the candidate source, and does not run the code. \cref{prompt:rationale} is the rationale it wrote for the first attempt of \cref{prompt:generation}, which reported ``optimal'' and matched the stored optimal objective.

\promptfile{Method-judge prompt}{judge.tex}{prompt:judge}

\promptfile{Judge rationale for Claude Sonnet 5, attempt 1 (the method-audit output)}{judge_rationale_sonnet5.tex}{prompt:rationale}

\subsection{Reflection}
\label{app:prompt-reflection}
The diagnosis prompt (\cref{prompt:reflection}) contains the inputs listed in \cref{tab:prompt-calls}. When the attempt is feasible and method-compliant but misses the 1\% target, the diagnosis prompt also contains the message in \cref{prompt:quality}, which withholds the optimal objective and the size of the gap. \cref{prompt:advice} is the advice written after the rejection in \cref{prompt:rationale}. The next generation call (\cref{prompt:revision}) repeats the generation instructions and appends the previous solver and its advice; the resulting attempt met all three criteria.

\promptfile{Reflection diagnosis prompt}{reflection_diagnosis.tex}{prompt:reflection}

\promptfile{Objective-quality message inserted into the diagnosis prompt}{quality_feedback.tex}{prompt:quality}

\promptfile{Reflection advice written by Claude Sonnet 5 after attempt 1 (the reflection advice file)}{reflection_advice_sonnet5.tex}{prompt:advice}

\promptfile{Revision generation prompt for reflection attempt 2}{revision_generation.tex}{prompt:revision}

\end{document}